\documentclass[11pt,a4paper]{article}
\pdfoutput=1
\usepackage[hyperref]{eacl2021}
\usepackage{times}
\usepackage{latexsym}
\usepackage{url}
\usepackage{amsmath}
\usepackage{amssymb}
\usepackage{tikz}
\usepackage{multirow}
\usepackage{adjustbox}
\usepackage{tabularx}
\usepackage{blindtext}
\usepackage{algorithm}
\usepackage[noend]{algorithmic}
\usepackage[font=small,skip=2pt]{caption}
\usetikzlibrary{positioning}
\usepackage{enumitem}
\usepackage{dsfont}
\usepackage[edges]{forest}
\usepackage{amsfonts}
\usepackage{booktabs}
\usepackage{siunitx}
\usepackage{makecell}
\usepackage{etoolbox}
\usepackage{cleveref}
\usepackage{soul}
\crefname{figure}{}{}
\crefname{table}{}{}
\usepackage{scrextend}

\newcommand\nl[1]{\emph{``#1"}}
\newcommand\sS{\mathcal{S}}
\newcommand\sC{\mathcal{C}}

\aclfinalcopy 

\definecolor{airforceblue}{rgb}{0.36, 0.54, 0.66}
\definecolor{battleshipgrey}{rgb}{0.52, 0.52, 0.51}
\definecolor{plot_blue}{rgb}{0.12, 0.46, 0.7}

\newcommand{\gtdone}[1]{#1}

\newcommand{\gtdeldone}[1]{}

\newcommand{\dhs}{HDS}
\newcommand{\mcdiv}{McDiv}
\newcommand{\mcdivh}{McDiv$_{\text{nuggets}}$}

\title{Evaluating the Evaluation of Diversity in Natural Language Generation}
\author{Guy Tevet$^{1,2}$ ~~~~~
Jonathan Berant$^{1,3}$ \\
\mbox{}\\
$^1$School of Computer Science, Tel-Aviv University \\
$^2$Department of Electrical Engineering, Tel-Aviv University \\
$^3$Allen Institute for AI \\
\small{\texttt{\{guytevet@mail,joberant@cs\}.tau.ac.il}}}

\begin{document}
\maketitle

\begin{abstract}

Despite growing interest in natural language generation (NLG) models that produce diverse outputs, there is currently no principled method for evaluating the diversity of an NLG system.
In this work, we propose a framework for evaluating diversity \emph{metrics}. 
The framework measures the correlation between a proposed diversity metric and a \emph{diversity parameter}, a single parameter that controls some aspect of diversity in generated text. For example, a diversity parameter might be a binary variable used to instruct crowdsourcing workers to generate text with either low or high content diversity.
We demonstrate the utility of our framework by: (a) establishing best practices for eliciting diversity judgments from humans, (b) showing that humans substantially outperform automatic metrics in estimating content diversity, and (c) demonstrating that existing methods for controlling diversity by tuning a ``decoding parameter" mostly affect form but not meaning.
Our framework can advance the understanding of different diversity metrics, an essential step on the road towards better NLG systems.
\end{abstract}

\section{Introduction}
\label{sec:intro}

\begin {figure}[!t]
\centering
\small
\resizebox {\columnwidth} {!} { %

\fbox{
\begin{tabularx}{\columnwidth}{l}

    {\resizebox {\columnwidth} {!} {
    \includegraphics{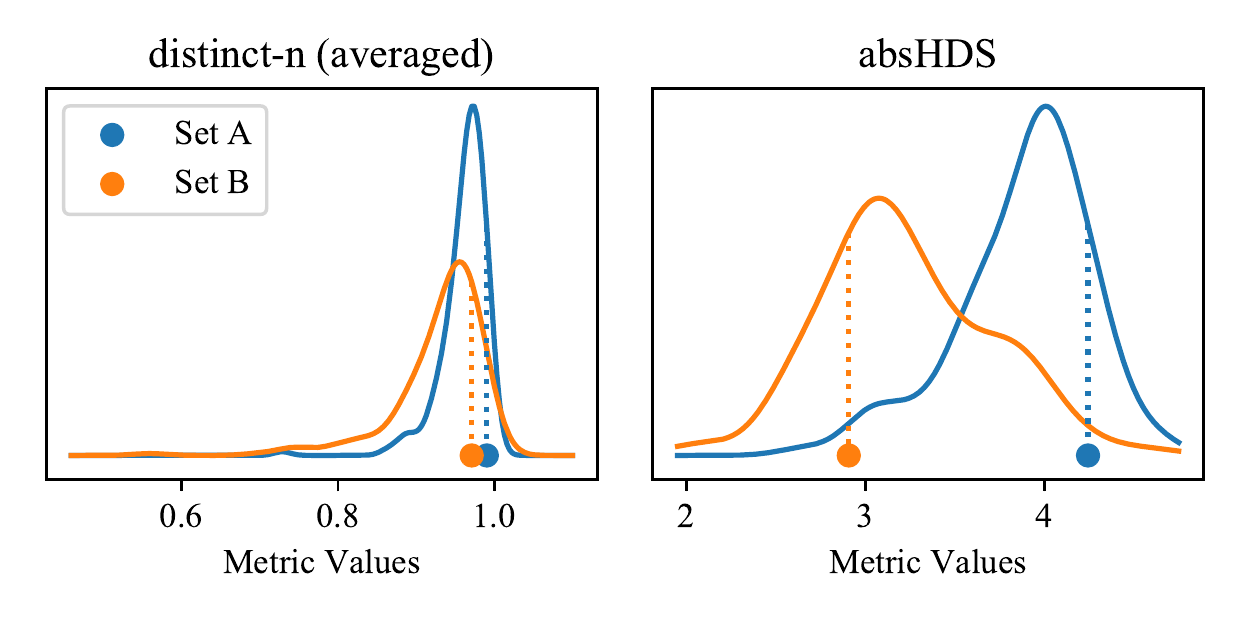}
    }} \\

    \multicolumn{1}{l}{\textbf{Question:} So what did I miss in the first 20 minutes?} \\
    \multicolumn{1}{l}{}  \\
    
    \textbf{\textcolor{blue}{Set A}} \\
    
    {\tiny$\bullet$} Pretty much everything.\\
    {\tiny$\bullet$} Nothing, really.\\
    {\tiny$\bullet$} You won't believe what happened!\\
    {\tiny$\bullet$} Why do you even care?\\
    {\tiny$\bullet$} What were you doing that was more important than this?\\
    \\
    
    \textbf{\textcolor{orange}{Set B}} \\
    
       {\tiny$\bullet$} Not much.\\
      
       {\tiny$\bullet$} It was pretty dull.\\
      
       {\tiny$\bullet$} Blah, you didn't miss anything.\\
      
       {\tiny$\bullet$} Not anything that important.\\
      
       {\tiny$\bullet$} Very little, it was uneventful.\\
    
    \multicolumn{1}{l}{} \\

\end{tabularx}
}
}
\caption{Diversity metric evaluation: we show two sets of responses to the same question, generated by crowdsourcing workers. While both sets are diverse in terms of \emph{form}, only set A is diverse in terms of \emph{content}.
Each graph presents the distribution over a diversity metric for sets with high content diversity (\textcolor{blue}{blue}) and low content diversity (\textcolor{orange}{orange}). 
Distributions are approximated over $200$ sets.
We observe that the human score metric (absDHS) separates the two distributions, while an n-gram based metric (distinct-n) fails, illustrating that it does not capture content diversity. The dotted lines correspond to the specific sets A and B presented above.}
\label{fig:fig1}
\end {figure}

An important desideratum of natural language generation (NLG) systems is to produce outputs that are not only \emph{correct}, but also \emph{diverse}.
For example, 
a dialog system \cite{adiwardana2020towards} should permit many responses for the prompt
\nl{How are you today?}. Similarly, we expect diverse responses in tasks such as story generation \cite{li2018generating}, question generation \cite{pan2019recent} and question answering \cite{fan2019eli5}.


Despite growing effort to produce more diverse models \cite{li2016simple,li2016diversity,holtzman2019curious,du2019boosting}, there is no standard evaluation metric for measuring diversity. Thus, different papers evaluate diversity differently (if at all), making it difficult to compare competing approaches \cite{hashimoto2019unifying}. Having a principled and consensual diversity evaluation metric is hence fundamental for the field of NLG.



A key challenge in developing diversity evaluation metrics, is the difficulty in determining their efficacy. 
Unlike metrics for evaluating the \emph{quality} of generated text, where one can measure correlation between a metric (such as BLEU \cite{papineni2002bleu}) and human judgement \cite{zhang2019bertscore,sagarkar2018quality}, it is unknown if humans can reliably estimate  diversity. 



In this paper, we propose a framework for evaluating diversity metrics (Figure~\ref{fig:test_flow}). We assume that a \emph{tester} (human or model) is generating sets of sentences, conditioned on some \emph{diversity parameter} that controls the diversity of the output sentences. We evaluate the diversity of the sentences using a proposed metric, and measure correlation between the metric and the diversity parameter. High correlation indicates that the metric captures how the diversity parameter affects the model output.

\gtdeldone{
We instantiate this framework with two tests. In the \emph{decoding test}, the tester is a neural generation model and the diversity parameter is a decoding parameter, such as softmax temperature \cite{ackley1985learning}.} 
\gtdeldone{
This parameter controls the skewness of the distribution in every generated token, and is known to affect model diversity \cite{holtzman2019curious,caccia2018language}. 
}
\gtdeldone{In the \emph{content test} (Figure~\ref{fig:fig1}), the tester is a \emph{human}, and the diversity parameter is a binary variable, where the human is instructed to generate sets of sentences with either \emph{high} or \emph{low} diversity \emph{in content}.
}

\gtdone{
We instantiate this framework with two tests. As a preliminary step, we introduce the \emph{decoding test}: the tester is a neural generation model and the diversity parameter is a decoding parameter, such as softmax temperature \cite{ackley1985learning}. This parameter controls the skewness of the distribution in every generated token, and has been shown to affect model diversity \cite{holtzman2019curious,caccia2018language}. Then, we turn the focus to \emph{content} diversity, introducing the \emph{content test} (Figure~\ref{fig:fig1}). Here, the tester is a \emph{human}, and the diversity parameter is a binary variable, where the human is instructed to generate sets of sentences with either \emph{high} or \emph{low} diversity \emph{in content}.
}

We evaluate three families of popular diversity metrics with these tests: (a) \emph{n-gram-based metrics} that estimate diversity based on surface patterns in a set of generated sentences, (b) \emph{neural metrics}: we propose a reduction from evaluating sentence similarity to evaluating diversity, then evaluate diversity using state-of-the-art sentence similarity models, and (c) \emph{human evaluation}: we explore multiple ways in which humans can be asked to estimate diversity, resulting in multiple Human Diversity Score (\dhs) variations.

\gtdeldone{(explicitly say this is not a form div test)
We find that n-gram-based metrics succeed in detecting diversity driven by decoding parameters, suggesting that such parameters mostly control the form of text rather than content.
}

\gtdone{Applying our tests leads to several findings: (i) In the \emph{decoding test}, n-gram-based metrics correlate well with decoding parameters, such as softmax temperature. While the goal of our framework is to evaluate diversity metrics, this result lets us reflect back on the tester itself and conclude that decoding parameters predominantly control the form of text rather than content. 
(ii) Conversely, n-gram-based metrics perform poorly in the \emph{content test}. While neural metrics outperform n-gram-based metrics, humans are substantially better than any automatic metric at detecting content diversity. This is illustrated in Figure~\ref{fig:fig1}, where a human clearly distinguishes between sets that have high (blue) and low (orange) content diversity, while n-gram-based metrics fail to do so.
}

Due to this gap, we construct a large dataset focused on \emph{content}-diversity metrics. We release the \textbf{M}etrics for \textbf{c}ontent \textbf{Div}ersity (\emph{\mcdiv{}}) benchmark, a challenge for research in diversity evaluation.

To conclude, our main contributions are:
\begin{itemize}[leftmargin=*,topsep=0pt,itemsep=0pt,parsep=0pt]
    \item A framework for evaluating diversity metrics.
    \gtdeldone{\item Tests instantiating this framework, measuring the sensitivity of metrics to content and form.}
    \item \gtdone{Tests instantiating this framework, measuring the sensitivity of metrics to diversity, with a focus on content diversity.}
    \item Best practices for obtaining diversity evaluations from crowdsourcing workers.
    \item Establishing that humans outperform current automatic metrics in detecting content diversity.
    \item The \mcdiv{} dataset - a benchmark for content diversity aware metrics.
    \item The collected data, test scores and code are publicly available,\footnote{\url{https://github.com/GuyTevet/diversity-eval}} and can be used to easily compare new diversity metrics to existing results in our framework.
\end{itemize}

\section{Background: Diversity Evaluation}
\label{sec:prelim}
Recently, interest in diversity has increased \cite{du2019boosting,holtzman2019curious}, resulting in multiple proposals for its evaluation. We describe recent approaches, highlighting the need for a standard way to evaluate metrics.

\paragraph{Perplexity} is the standard metric in language modeling, measuring the proximity of a language model (LM), $P_{\text{LM}}$, to the true distribution, $P_{\text{ref}}$, by approximating the cross-entropy $H(P_{\text{ref}}, P_{\text{LM}})$ with held-out data from $P_{\text{ref}}$. Thus, perplexity captures to some extent diversity. For example, a dialog model that puts all probability mass on the output \nl{I don't know} for any given context will obtain infinite perplexity once it encounters any other response. This property makes perplexity popular in LM-based NLG models, and often it is the only reported measure for diversity \cite{lewis2017deal,fan2018hierarchical,wang2019paperrobot,li2019generating}.

However, perplexity does not purely measure diversity, and high perplexity does not entail low diversity. For example, a LM with a uniform distribution over the vocabulary for each decoded token has high diversity, but its perplexity will be extremely high, due to its low \emph{quality}. 
Moreover, perplexity evaluates a LM, while the diversity of a NLG system is also strongly affected by the decoding procedure. For example, \textit{Top-k} and \emph{nucleus sampling} are popular decoding schemes that trade-off quality and diversity by ignoring some of the LM probability mass \cite{holtzman2019curious}.


Last, some NLG models, such as Generative Adversarial Networks (GANs) \cite{yu2017seqgan} are not language models. While one can approximate perplexity for such models \cite{tevet2019evaluating}, ideally, a metric should not be tied to a model.

\paragraph{N-gram-based metrics} A popular metric is \emph{distinct n-grams} \cite{li-etal-2016-diversity}, which computes the proportion of unique n-grams out of the total number of n-grams in a set of generated sentences.
\newcite{duvsek2020evaluating} calculated \emph{Shannon entropy} 
\cite{manning1999foundations} 
based on different n-grams as a measure of lexical diversity.
\emph{Self-BLEU} \cite{zhu2018texygen,shu2019generating} measures the BLEU score of a generated sentence with respect to another generated sentence (rather than a gold reference). High average Self-BLEU indicates high similarity between generated sentences and low diversity. In \S\ref{sec:sim_red} we expand this idea and suggest a reduction from any similarity metric to a diversity metric. By design, n-gram based metrics are sensitive to diversity in the \emph{form} of language, rather than its meaning.

\paragraph{Embedding-based metrics} A new line of metrics suggests to embed generated sentences in latent space, then evaluate them in this space. \newcite{du2019boosting} suggest to cluster the embedded sentences with k-means, then use its inertia as a measure for diversity. Recently, \newcite{lai2020diversity} suggested to consider the volume induced
by the embedded sentences as a diversity metric.

\paragraph{Human evaluation} 
\newcite{yang2019enhancing} asked humans to evaluate the internal diversity of a generated essay. \newcite{ghandeharioun2019approximating} let crowdsourcing workers interact with a dialog chat-bot, then asked them to evaluate the diversity of a single conversation. In contrast, this paper focuses on the diversity of different responses given a context, as in \newcite{zhang2019syntax}.



To conclude, increasing interest in diversity resulted in multiple proposed diversity metrics. However, there is no consensus on how to evaluate diversity and what each metric actually measures.

\section{Evaluating Diversity Metrics}
\label{sec:d_benchmark}

\begin {figure}[!t]
\centering
\resizebox {0.8\columnwidth} {!} { %

\begin{tikzpicture}[node distance=1.5cm,
every node/.style={fill=white}, align=center]

\tikzset{
	base/.style = {rectangle, rounded corners,
		minimum width=3.5cm, minimum height=1.4cm,
		text centered}, 
	io/.style = {base},
	block/.style = {base, draw=black},
	example/.style = {text=blue, align=left},
	line/.style={
    draw,rounded corners=3mm, -latex,
    }
}

\node (div_param) [io] {Diversity Parameter\\$d$};
\node (context) [io, right of=div_param, right=2cm] {\nl{How are you today?}\\ $c$};
\node (tester) [block, below of=div_param, right=1cm] {Tester  \includegraphics[height=10pt]{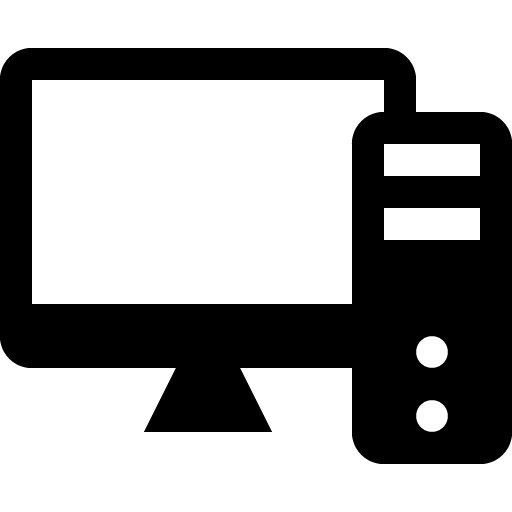} / \includegraphics[height=10pt]{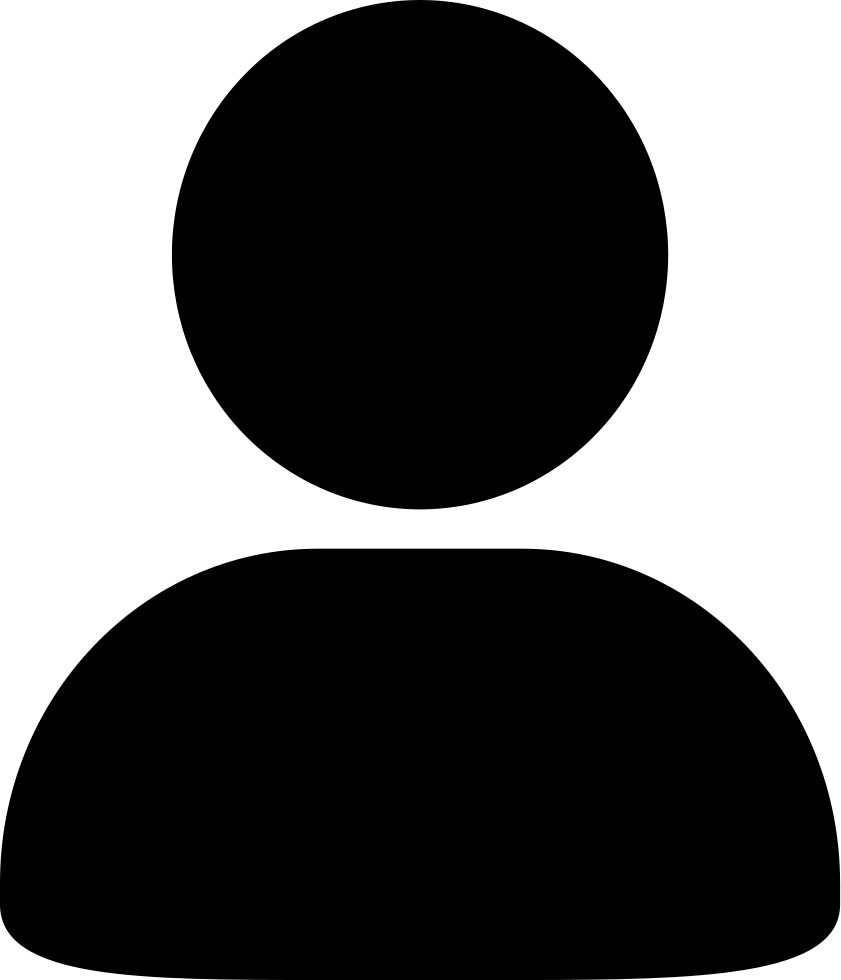}\\ $G_{d}(c)$};
\node (metric) [block, below of=tester, below=1.4cm] {Diversity Metric\\ $m_{\text{div}}(\sS_{c,d})$};
\node (rho) [block, below of=metric, below=0.1cm] {Test Score\\ $\rho(m_{\text{div}}, d)$};




\path[line]     (div_param.290) |-  (tester);
\path[line]     (context) |- (tester);
\path [line]     (tester) -- node [right, align=left]{\nl{Very good!}\\ \nl{Fine thank you.} \\ \nl{Couldn't be better.}}(metric);
\path[line]     (metric) -- node [right, align=center]{}(rho);
\path[line]     (div_param.250) |- (rho);

\end{tikzpicture}

}
\caption{An overview of our diversity metrics evaluation framework. The tester (machine or human) generates a response set ($\sS_{c,d}$) given a diversity parameter ($d$) and a context ($c$). The test score of a metric $m_\text{div}$ is the correlation between the metric score for $\sS_{c,d}$ and $d$.}
\label{fig:test_flow}
\end {figure}

We now describe our framework for evaluating diversity metrics. Diversity has many facets: 
for instance, a set of sentences can be diverse in terms of their \emph{content}, while another may have similar content, but diverse \emph{form} (Figure~\ref{fig:fig1}). Our framework provides a way to evaluate metrics for different aspects of diversity under moderate assumptions.

We define a diversity metric $m_{\text{div}}(\sS_c) \in \mathbb{R}$ as a function that takes a set of generated responses $\sS_c$ 
as an input, and outputs a diversity score. Each response $s \in \sS_c$ is generated for the same input context $c$, hence $\sS_c$ is a sample from a generative distribution $P_{\text{gen}}(s \mid c)$. The overall diversity score of a generative model 
can be obtained by averaging $m_{\text{div}}$ over sets $\sS_c$ sampled from the model given multiple contexts $c \in \sC$.



To evaluate $m_{\text{div}}(\cdot)$, we assume access to some deterministic \emph{diversity parameter} $d$ that controls an aspect of diversity in $\sS_c$. We test the relation between $m_{\text{div}}$ and the parameter $d$. By varying $d$ and measuring $m_{\text{div}}$, we can compute the correlation $\rho$ between $m_{\text{div}}$ and an aspect of diversity represented by $d$.
Because our goal is to have metrics that \emph{rank} the diversity of generated texts,
 we use Spearman's $\rho$ rank correlation 
 as our test score.
Figure~\ref{fig:test_flow} illustrates the flow of a test in our framework.


In practice, to control the diversity level of $\sS_c$ using $d$, we use \emph{a tester}: a generative model that takes a context $c$ and a diversity parameter $d$ as input, and outputs a response set $\sS_{c,d}$. We stress that the tester can be either a neural model or a human. A good tester should reliably represent the diversity level quantified by $d$.

As a hypothetical example, $c$ can be a movie name and $d$ represent \emph{sentiment diversity}, that is,  the number of different sentiments in a collection of reviews $\sS_c$. A human tester can observe $c$ and $d$, and produce reviews 
accordingly
(such data can be easily mined from IMDB). A collection of such $(d, \sS_{c,d})$ makes a test, in which the correlation between  $m_{\text{div}}(\sS_{c,d})$ and $d$ measures the sensitivity of $m_{\text{div}}$ to sentiment diversity.

\gtdeldone{ (moved to a footnote)
We note that perplexity cannot be evaluated as a diversity metric in our framework, because it requires a sample from $P_\text{ref}$, while we assume a response set sampled from $P_\text{gen}$.
}





We now describe two tests that instantiate this framework, roughly corresponding to the two main aspects of diversity: form diversity and content diversity.

\subsection{Decoding Test}
\label{sec:test1}

The diversity of a NLG system  constructed from a LM depends on both the LM but also the decoding algorithm on top of it.
For example, \textit{beam search} approximates the most probable output, and dramatically reduces diversity.
Conversely, sampling from the LM leads to high diversity, but low quality output \cite{holtzman2019curious}. 

A popular method to control diversity in NLG systems is to vary some decoding parameter. 
Variations include (a) \textit{softmax temperature} \cite{ackley1985learning}, where a parameter $\tau$ controls the skewness of the softmax distribution at each step, (b) \textit{Nucleus (Top-$p$) sampling} \cite{holtzman2019curious}, where one samples at each step from the minimal set of most probable tokens whose cumulative probability is at least $p$, and (c) \textit{Top-$k$ sampling}, 
which samples from 
the top-$k$ most probable 
tokens at each step.
All methods skew the LM distribution in a way that avoids low-probability tokens and leads to higher quality \cite{holtzman2019curious}, providing a \emph{decoding parameter} that trades off quality and diversity \cite{caccia2018language}.

In the decoding test (\emph{decTest}), we define the \emph{tester} to be a LM, such as GPT-2 \cite{radford2019language}, and the diversity parameter $d$ to be a decoding parameter such as temperature. We check how different diversity metrics $m_\text{div}$ correlate with decoding parameters. This can shed light on the quality of the metrics, but also on how decoding parameters affect the output of a NLG system.
\gtdone{
The decoding test uses automatically-generated data that is cheap to produce, and decoding parameters that are well-known to control diversity. Thus, we view this test as a warm-up test to explore the strengths of our framework.
}




\subsection{Content Test}
\label{sec:test2}

In the content test (\emph{conTest}), our goal is to evaluate how different diversity metrics capture the notion of \emph{content diversity}.
Measuring content diversity requires deep understanding of the semantics of responses in $\sS_c$.

To isolate \textit{content} from \textit{form} diversity, we aim to generate response sets with a similar level of form diversity, but where the level of content diversity is controlled by the diversity parameter $d$. Thus, we use crowdsourcing workers as testers, and a binary parameter $d \in \{0, 1\}$, corresponding to low or high content diversity.
A worker observes a context $c$ and produces a set of responses $\sS_c$ based on the value of $d$. We encourage workers to use different words and phrases in different responses regardless of the value of $d$, such that form diversity is high in all examples. Examples from this data are in Figure~\ref{fig:fig1} and 
Appendix \ref{app:data_samples}.

In \S\ref{sec:experiments}, we will focus on whether automatic diversity metrics can perform as well as humans on the task of estimating content diversity. 
\section{Human Diversity Score}
\label{sec:dhs}

One of the core questions we tackle is:
\emph{Can humans evaluate diversity reliably?}
Although a few papers \cite{ghandeharioun2019approximating,yang2019enhancing,zhang2019syntax} asked humans to evaluate diversity, to the best of our knowledge no work thoroughly investigated this question. The importance of this question is clear when comparing to quality evaluation. There, human judgment is the gold standard, and automatic quality metrics are established by showing high correlation with human score. Thus, understanding if humans can judge diversity is important for improving diversity metrics.
We use crowdsourcing workers\footnote{Native English speakers, for more details see 
Appendix~\ref{app:dhs_qs}.
} to compute a human diversity score:
we show workers a context followed by a set of responses, and ask them to rate the diversity of the set.


To establish best practices, we experiment with multiple variations of \dhs{} (detailed in \S\ref{sec:human_metrics}), asking humans to rate the diversity of a response set, and evaluating each practice with our framework. We focus on the following questions:
\begin{itemize}[leftmargin=*,topsep=0pt,itemsep=0pt,parsep=0pt]
  \item Should humans rate \textit{diversity} of a set or similarity between pairs in the set, from which diversity can be inferred? \emph{(tl;dr: diversity)}
  \item Can humans evaluate different aspects of diversity well? \emph{(tl;dr: not effectively)}
  \gtdone{\item Should humans rate the \emph{absolute} diversity score of a set of sentences or \emph{rank} whether one set is more diverse than another? Here, we did not reach a conclusive result, and describe this experiment in the 
  Appendix \ref{app:full_res}.}
\end{itemize}

As a preliminary step, we conducted pilot experiments among a group of NLP graduate students. The main insights were: (a) humans are biased by quality:  if a generated set has high diversity but low quality, humans will rate diversity low. To neutralize this, we explicitly ask workers to evaluate the quality of one of the responses in the set $\sS_c$, and then instruct them to ignore quality in diversity questions; (b) To make sure a worker reads the context $c$, we ask them to generate a sentence $s$ before they rate diversity; (c) It is difficult for workers to evaluate the diversity of a set with more than 10 responses. Our crowdsourcing tasks are provided in 
Appendix~\ref{app:dhs_qs}.
 

\section{Diversity to Similarity Reduction}
\label{sec:sim_red}

We expand the idea from  \newcite{zhu2018texygen} and suggest a method to construct a diversity metric from any 2-sentence similarity metric. Given $m_{\text{sim}}(s_1, s_2) \in \mathbb{R}$, a symmetric similarity metric that gets a pair of input sentences $(s_1, s_2)$ and returns a similarity score, we can define a diversity metric $\tilde{m}_{\text{div}}$ as the negation of the mean similarity score across all (unordered) pairs of $\sS_c$:
\[ \tilde{m}_{\text{div}} (\sS_c) = - \frac{1}{\binom{|\sS_c|}{2}} \sum_{s_i, s_j \in \sS_c, i > j} m_{\text{sim}}(s_i, s_j). \]
This reduction allows us to easily define new diversity metrics based on past work on sentence similarity \cite{gomaa2013survey,devlin2019bert,zhang2019bertscore,reimers2019sentence}. In \S\ref{sec:experiments} we show that both n-gram-based similarity metrics and neural semantic similarity metrics provide useful diversity metrics. 


\section{Experiments}
\label{sec:experiments}


\subsection{NLG Tasks}
\label{sec:tasks}

We apply our evaluation procedure on three different English NLG tasks that require diversity.
\begin{itemize}[leftmargin=*,topsep=0pt,itemsep=0pt,parsep=0pt]
    \item \textbf{Story completion (\textit{storyGen}); } We use the  ROC Stories dataset \cite{mostafazadeh-etal-2016-corpus}, in which the context $c$ is the first four sentences of a story, and the response $s$ is a single sentence that ends the story. We use the contexts $\sC$ from this data and generate response sets $\sS_c$ for each context using our testers.
    The long contexts characterizing this data narrow down the space of possible responses, making this a ``low-entropy" generation task, where the output is constrained, but diversity is still essential.
    \item \textbf{Dialog response generation (\textit{respGen}); } A comment-response pairs dataset extracted from the website \url{reddit.com}
    and pre-processed by \newcite{hashimoto2019unifying}. 
    We use the comments from their data as contexts $\sC$  and generate response sets $\sS_c$ for each context using our testers. 
    Since comments are single sentences the response is less constrained, making this a ``medium-entropy" generation task.
    \item \textbf{3-words prompt completion (\textit{promptGen}); } Contexts $\sC$ are 3-words prompts, extracted from the Cornell Movie-Dialogs Corpus \cite{Danescu-Niculescu-Mizil+Lee:11a}
     by taking the first three words from each original context. The response sets $\sS_c$ are completions of the prompts, generated by our testers. 
    This context provides minimal constraints, making this a ``high-entropy" generation task.
\end{itemize}
Samples of the contexts extracted for each task, along with generated response sets, are presented in 
Appendix~\ref{app:data_samples}.
We intentionally avoid NLG tasks where diversity is not necessarily desired, such as summarization and machine translation.


\subsection{Evaluated Metrics}
\label{sec:auto_metrics}

\noindent
\textbf{N-gram-based metrics}
We evaluate distinct n-grams (\emph{distinct-n}), as described in \S\ref{sec:prelim}. We also evaluate n-grams cosine similarity (\emph{cos-sim}): a similarity measure computing the cosine between the vectors representing two sentences, where each vector is a count vector over the n-grams that appear in the response. We use the reduction from \S\ref{sec:sim_red} to convert this to a diversity measure.
In both metrics, rather than choosing the order of the n-grams, we average over $n \in \{1,\dots, 5\}$, which we found to outperform any single choice of $n$.


\noindent
\textbf{Neural metrics}
We exploit existing BERT-based models \cite{devlin2019bert} fine-tuned for estimating similarity between two sentences (applying the reduction from \S\ref{sec:sim_red}).


\noindent
\textit{BERT-STS}; A BERT model fine-tuned on Semantic Textual Similarity \cite{cer2017semeval}: a collection of sentence pairs annotated with scores from 1-5 denoting their semantic similarity.\footnote{ \url{https://github.com/swen128/bert-sts}}


\noindent
\textit{BERT-Score} \cite{zhang2019bertscore}; Originally a quality metric, \textit{BERT-Score} uses BERT's embeddings to measure similarity between two sentences.
We used \texttt{RoBERTa-large} \cite{liu2019roberta}, as suggested by the authors.\footnote{\url{https://github.com/Tiiiger/bert_score}}

\noindent
Sentence-BERT (\textit{sent-BERT}) \cite{reimers2019sentence} is a sentence-level embedding model based on BERT. We use the cosine similarity between the embeddings of two responses as a similarity metric.
In our experiments we used \texttt{bert-large-nli-stsb-mean-tokens}.\footnote{\url{https://github.com/UKPLab/sentence-transformers}}

\noindent
\textbf{Human Metrics}
\label{sec:human_metrics}
We examine four methods for evaluating diversity with humans (see \S\ref{sec:dhs}), to investigate best practices for obtaining diversity judgment from humans.
In all metrics (except ranking), ratings are from 5 (highest diversity/similarity) to 1 (lowest). The original tasks presented to workers are in 
Appendix~\ref{app:dhs_qs}.

\noindent
Absolute HDS (\textit{absHDS});
Given a context $c$ and a set of generated responses $\sS_c$, rate the level of diversity of $\sS_c$.


\noindent
Ranking HDS (\textit{rnkHDS}); Given a context $c$ and \underline{two} sets $\sS_{c, d_1}, \sS_{c, d_2}$ generated with different values of the diversity parameter $d$, rate which set is more diverse. Since this metric did not clearly outperform \textit{absHDS}, we provide results in 
Appendix \ref{app:full_res} only.

\noindent
Similarity HDS (\textit{simHDS});
Given a context $c$ and a set of generated responses $\sS_c$, rate the similarity of each two sentences in  $\sS_c$, and then apply the reduction from \S\ref{sec:sim_red}.


\noindent
Aspect HDS (\textit{aspHDS});
Identical to \textit{absHDS}, except we explicitly ask about a specific aspect of diversity, namely \emph{form} and \emph{content}.\footnote{\gtdone{We note that perplexity cannot be evaluated as a diversity metric in our framework, because it requires a sample from $P_\text{ref}$, while we assume a response set sampled from $P_\text{gen}$.}}

\begin {figure*}[!t]
\makebox[\textwidth][c]{\includegraphics[width=0.9\textwidth]{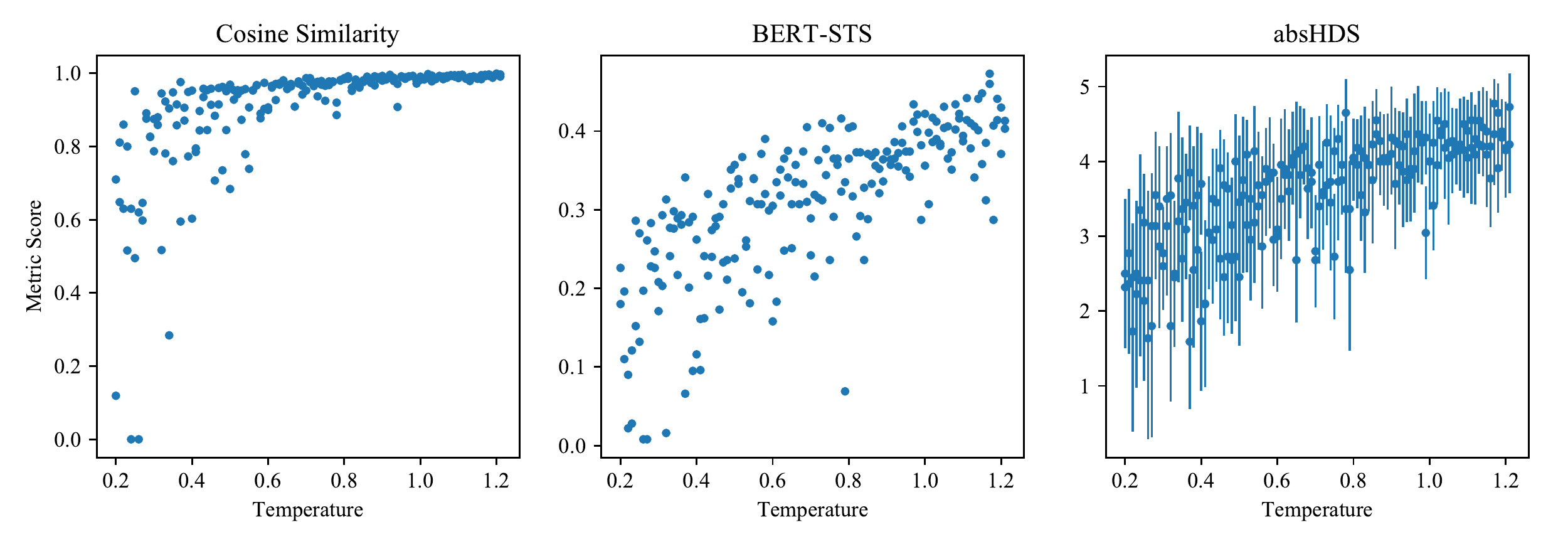}}
\caption{ \emph{decTest}: Scatter plot of n-gram-based (cosine similarity), neural (BERT-STS) and human (abs\dhs{}) metrics as a function of temperature for \textit{respGen}. Each point corresponds to a single generated set. Error bars of \dhs{} represent the standard deviation over 10 annotator ratings.}
\label{figure:test1}
\end {figure*}

\subsection{Decoding Test}
\label{sec:exp_test1}

\begin {table}
\centering
\footnotesize
\footnotesize

\begin{tabular}{l} 
    \toprule
    \textbf{Context}\\
    \midrule
    
    \makecell[l]
    {
    \textbf{Fire next door.}\\
    John woke up smelling like something was burning.\\
    He went outside.
    He saw the fire next door.\\
    He called the authorities.} \\
    \midrule
    \textbf{Response set ($\tau = 0.25$)}\\
    \midrule
    \makecell[l]
    {
    {\tiny$\bullet$} It was a minor fire and they put it out.\\
    {\tiny$\bullet$} It was a fire.\\
    {\tiny$\bullet$} It was a fire.\\
    {\tiny$\bullet$} It was a fire.\\
    {\tiny$\bullet$} It was a fire.\\
    } \\
    \midrule
    \textbf{Response set ($\tau = 0.8$)}\\
    \midrule
    \makecell[l]
    {
    {\tiny$\bullet$} They arrived and put out the fire. \\
    {\tiny$\bullet$} It was a fire. \\
    {\tiny$\bullet$} It was a fire. \\
    {\tiny$\bullet$} It turned out to be a fire. \\
    {\tiny$\bullet$} It was a minor fire night. \\
    } \\
    \midrule
    \textbf{Response set ($\tau = 1.1$)}\\
    \midrule
    \makecell[l]
    {
    {\tiny$\bullet$} It turned out to be a mechanic.\\
    {\tiny$\bullet$} Before the fire was put out it was a fire.\\
    {\tiny$\bullet$} It was a fire.\\
    {\tiny$\bullet$} They co-worker matter how bad the fire was.\\
    {\tiny$\bullet$} Several shells, the fire department came just in time.\\
    }\\

    \bottomrule
\end{tabular}
\caption{An example of the effect of \emph{temperature} on the response set $\sS_c$ for a context $c$ from ROC Stories.}
\label{table:temp_example}
\end {table}

In decTest we measure the correlation between diversity metrics ($m_{\text{div}}$) and the \textit{softmax temperature} decoding parameter ($d$). The tester generating the response sets ($\sS_c$) is a neural NLG model.

\paragraph{Data and settings}
\label{sec:exp_test1_data}
For each task, we generated sets of $10$ responses per context, using a linear temperature sweep with $100$ values in the range $[0.2, 1.2]$  \cite{caccia2018language}. We generated 1K sets in total for each of 1K contexts ($10$ per temperature) and evaluated $200$ ($2$ random sets per temperature).  
For automatic metrics, we repeat this 100 times (randomly sampling $200$ out of $1K$ sets each time), to present the mean and standard deviation. 
\dhs{} metrics are computed over one experiment of $200$ sets, due to their high cost.

Data for \textit{storyGen} and \textit{respGen} was generated by the MASS model \cite{song2019mass}, fine-tuned on each dataset. Data for \textit{promptGen} was generated by GPT-2-\emph{large} \cite{radford2019language} without fine-tuning. We provide examples for how story endings change as a function of temperature in Table~\ref{table:temp_example}. Examples for all tasks \gtdone{along with additional reproducibility details} are in the
Appendix~\ref{app:data_samples}.
For each \dhs{} metric, we collected 10 ratings per query from Amazon Mechanical Turk (AMT) workers. 
While abs\dhs{} demands one query per response set, in order to perform sim\dhs{} at a reasonable cost, we chose $|\sS_c|=5$,
resulting in $\binom{5}{2} = 10$ crowdsourcing queries instead of $\binom{10}{2} = 45$ per set.
\gtdone{We evaluate sim\dhs{} only for \textit{respGen} due to the metric's high cost and low performance.}

\paragraph{Results}
Table~\ref{table:test1_abs} presents results of abs\dhs{}, sim\dhs{}, and all automatic metrics. In general, n-gram based metrics capture the diversity induced by a temperature sweep, beating \dhs{} and neural metrics.
Figure~\ref{figure:test1} provides a more detailed analysis. Each point represents a single set of responses generated at some temperature. While rank correlation for cosine similarity is high, it is far from linear and reaches high values even at low temperatures, scoring $0.6$ Pearson correlation. Conversely, the correlation for BERT-STS and abs\dhs{} is more linear, scoring $0.75$ and $0.77$ Pearson correlation respectively. Thus, Pearson and Spearman correlations disagree on the quality of the different metrics in this case.

While our framework is meant to evaluate diversity metrics, the results of the test let us reflect on the decoding parameters themselves.
This result shows that humans perform worse than automatic metrics in this experimental setup, hinting that temperature mostly controls superficial changes to the generated text. Additionally, sim\dhs{} performs worse than abs\dhs{} 
although it is 3x more expensive, showing that rating the entire set rather than averaging over pairs is useful.



\begin {table}
\centering
\footnotesize

\begin{tabular}{lccc} \toprule
      & \textbf{storyGen} & \textbf{respGen} & \textbf{promptGen} \\
    \midrule
     distinct-n  & \textbf{0.76} (0.03) & \textbf{0.89} (0.01) & \textbf{0.91} (0.01) \\
     cos-sim & 0.71 (0.04) & \textbf{0.89} (0.01) & 0.87 (0.02) \\
    \midrule
     BERT-STS  & 0.64 (0.04)   & 0.81 (0.02) &  0.84 (0.02) \\
      sent-BERT  & 0.65 (0.03)   & 0.80 (0.02) & 0.74 (0.03) \\
     BERT-score  & 0.69 (0.04)   & 0.87 (0.01) & 0.88 (0.02) \\
    \midrule
    abs\dhs  & 0.69   & 0.81 & 0.79 \\
     sim\dhs  & -   & 0.74 & - \\
 \bottomrule
\end{tabular}

\caption{\emph{decTest} results: Spearman's $\rho$ correlation between temperature and each metric score (mean and standard deviation). 
\textit{sim\dhs} was tested only on \emph{respGen}. }
\label{table:test1_abs}
\end {table}

\paragraph{Other decoding parameters}

To compare the robustness of our conclusions to other decoding parameters, we repeat it with two additional decoding methods: (a)
in \textit{Nucleus (Top-$p$) sampling} we swept linearly over 100 values of $p$ in the range $[0.1, 1.0]$; (b) In \textit{Top-$k$} sampling we swept $k$ in logarithmic scale over 100 values in the range $[1, 30K]$ and present the correlation between the metrics and $\log_{10}(k)$.
While softmax temperature enables skewing $P_{\text{LM}}$ to a more diverse $P_{\text{gen}}$ using $\tau > 1$, both Top-$p$ and Top-$k$ enable only skewing $P_{\text{LM}}$ to a more sharp (hence less diverse) $P_{\text{gen}}$.  


Table \ref{table:test1_d_params} presents results for all automatic metrics using the three decoding methods over \emph{promptGen}. Results for other tasks are in
Appendix~\ref{app:full_res}. 
\gtdeldone{Although the correlation in Top-$k$ is significantly lower, and the variance is higher, all three decoding methods reflect a similar ordering between the metrics.}
\gtdone{We find that Top-$p$ correlates well with temperature along all three generation tasks, whereas Top-$k$ does not correlate with any of them.}

\begin {table}[!t]
\centering
\resizebox {\columnwidth} {!} { %
\footnotesize

\begin{tabular}{lccc} \toprule
     & \textbf{Temperature} & \textbf{Top-p} & \textbf{Top-k} \\
    \midrule
    distinct-n  & \textbf{0.91} (0.01) & \textbf{0.84} (0.02) & \textbf{0.61} (0.05) \\
    cos-sim & 0.87 (0.02) & 0.78 (0.03) & 0.48 (0.05) \\
    \midrule
    BERT-STS & 0.84 (0.02)   & 0.74 (0.03) & 0.55 (0.05) \\
    sent-BERT  & 0.74 (0.03)   & 0.63 (0.05) & 0.51 (0.05) \\
    BERT-score  & 0.88 (0.02)   & 0.77 (0.03) & 0.57 (0.05) \\

 \bottomrule
\end{tabular}

}
\caption{ \emph{decTest} results for different decoding parameters: Spearman's $\rho$ (mean and standard deviation) of automatic metrics for \emph{promptGen}.}
\label{table:test1_d_params}
\end {table}

\subsection{Content Test}
\label{sec:exp_test2}

In conTest, we measure the correlation between diversity metrics ($m_{\text{div}}$) and content diversity, represented by a binary parameter $d \in \{0,1\}$. The testers are AMT workers, guided to create sets with high level of \emph{form} diversity and high or low \emph{content} diversity according to $d$.

\paragraph{Data and settings}

\begin {figure*}[!t]
\makebox[\textwidth][c]{\includegraphics[width=0.9\textwidth]{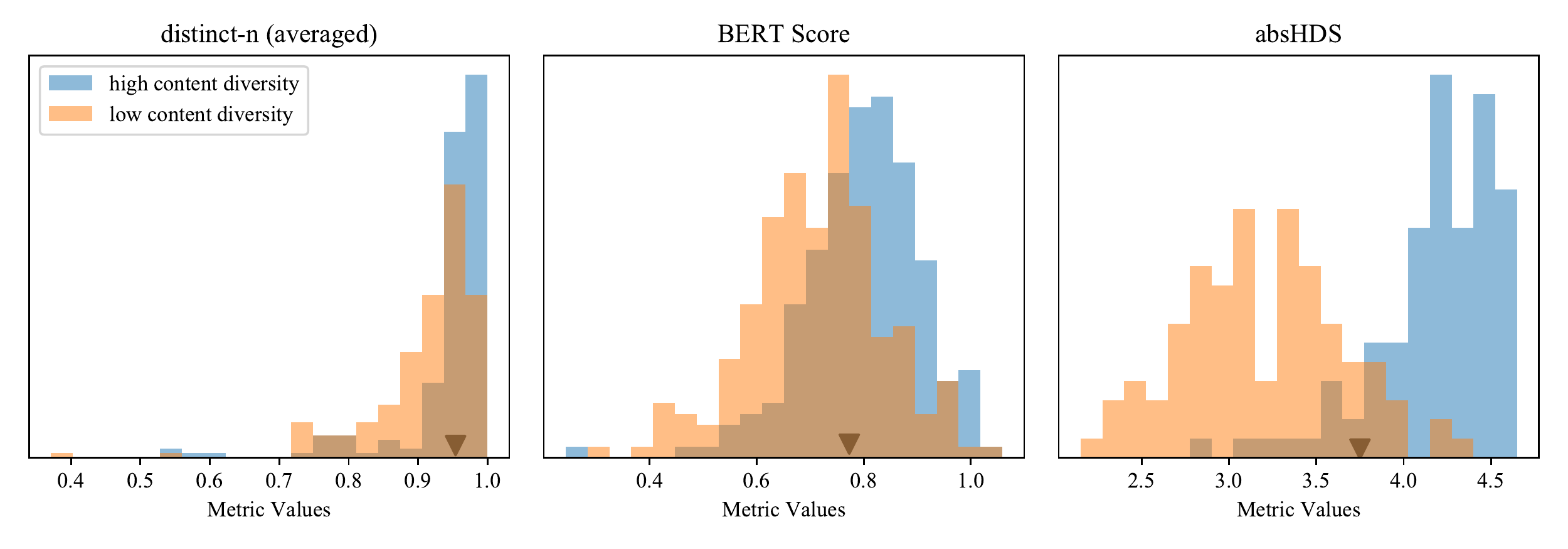}}
\caption{ \emph{conTest}: histograms of metric values of n-gram (distinct n-grams), neural (BERT-Score) and human (abs\dhs) metrics for \emph{promptGen}. The \textbf{\textcolor{orange}{orange}} histogram represents the distribution of the \textit{low content diversity class}, the \textbf{\textcolor{blue}{blue}} histogram represents the distribution of the \textit{high content diversity class} and \textbf{\textcolor{brown}{brown}} is the intersection between the two. Pointing down triangles represent the threshold $\eta$ of the optimal classifiers. The histograms show how each metric separates the two classes.}
\label{figure:test2}
\end {figure*}

For each task, we collected 200 sets of 5 responses each (100 sets per class). For high content diversity class, we asked workers to give 5 responses per context, with as different content and structure as possible. Then we asked the same workers to choose a single response they wrote, and rephrase it 5 times such that the original content will be preserved, while changing the form -- this set is used for the low content diversity class. A sample from this data is in Figure \ref{fig:fig1} and more samples in 
Appendix~\ref{app:data_samples}.
For each \dhs{} metric, we collected 10 ratings from crowdsourcing workers, different than the ones who composed the sets. 

\paragraph{Results}
In addition to Spearman's $\rho$, we report the optimal single-threshold classifier accuracy (OCA), i.e., the best achievable accuracy in predicting the class of a response set (high or low content diversity) for any threshold $\eta$ on $m_\text{div}$, such that if $m_\text{div}(\sS_c) > \eta$ the classifier predicts \emph{high diversity}, and otherwise predicts \emph{low diversity}.

Table~\ref{table:test2} shows the results. N-gram-based metrics perform poorly, indicating they do not measure content diversity well.
Neural models perform better than n-gram-based metrics (especially sent-BERT), but there is still a clear gap between automatic metrics and humans. Figure~\ref{figure:test2} illustrates the typical distributions of n-gram, neural and human metrics. Clearly, \dhs{} separates high and low \emph{content} diversity better than neural metrics. In addition, n-gram-based metrics saturate both classes to near maximal values, similarly to decTest.

Since conTest isolates content diversity, we used asp\dhs{} to directly rate content and form diversity. \emph{Content} asp\dhs{} gets similar scores to \textit{abs\dhs}, suggesting little gain in asking directly on the tested aspect. \emph{Form} asp\dhs{} gets low scores compared to abs\dhs{}, validating that the form diversity of the two classes is similar.


\begin {table}[!t]
\centering
\resizebox {\columnwidth} {!} { %
\footnotesize

\begin{tabular}{l|cc|cc|cc} 
    \toprule
     & \multicolumn{2}{c}{\textbf{storyGen}} & \multicolumn{2}{c}{\textbf{respGen}} & \multicolumn{2}{c}{\textbf{promptGen}} \\
    & $\rho$ & OCA & $\rho$ & OCA & $\rho$ & OCA\\
    \midrule
    distinct-n  & 0.57 & 0.77 & 0.34 & 0.67 & 0.33 & 0.68\\

    cos-sim & 0.56 & 0.77 & 0.33 & 0.66 & 0.36 & 0.67\\
    \midrule
    BERT-STS  & 0.6 & 0.78 & 0.46 & 0.72 & 0.65 & 0.82\\
    sent-BERT  & 0.77 & 0.90 & 0.59 & 0.79 & 0.68 & 0.81\\
    BERT-score  & 0.59 & 0.77 & 0.49 & 0.74 & 0.4 & 0.69\\

    \midrule
    abs\dhs  & \textbf{0.85} & \textbf{0.95} & 0.63 & 0.81 & \textbf{0.78} & \textbf{0.89}\\

    asp\dhs$_{\text{form}}$  & 0.35 & 0.65 & 0.56 & 0.79 & 0.4 & 0.68\\

    asp\dhs$_{\text{content}}$  & 0.84 & 0.94 & \textbf{0.67} & \textbf{0.83} & 0.75 & 0.88\\

 \bottomrule
\end{tabular}
}
\caption{\emph{conTest} results: Spearman's ($\rho$) correlation between a set's class and each metric score. 
}
\label{table:test2}
\end {table}

\paragraph{Content Diversity Benchmark}
\label{sec:mcdiv}
We construct the \textbf{M}etrics for \textbf{c}ontent \textbf{Div}ersity (\emph{\mcdiv{}}) benchmark, focusing on metrics for content diversity. \mcdiv{} is a dataset containing $6K$ $\{c,\sS_c\}$ pairs, ($2K$ for each storyGen, respGen and promptGen) collected as described in this section. 
\mcdiv{} contains a subset of $3K$ examples, termed \emph{\mcdivh{}}, in which \emph{form} diversity was neutralized, providing a difficult meta-evaluation challenge. \mcdivh{} was sampled to ensure that the correlation of \emph{distinct-n} (a form diversity metric) is zero over this subset.
Applying conTest over the data shows that n-gram based metrics obtain near-zero values on \mcdivh{} as expected, and all neural metrics perform substantially worse on \mcdivh{} than on \mcdiv{}.
On conTest, we obtain abs\dhs{} annotations for more than 200 random samples from \mcdivh{} and obtain 0.7 Spearman's $\rho$ for the respGen task, substantially higher than the best performing neural metric (sent-BERT) score at 0.6.
Details and conTest results can be found in 
Appendix \ref{app:full_res}.

\begin {figure}
\centering
\resizebox {\columnwidth} {!}{\includegraphics{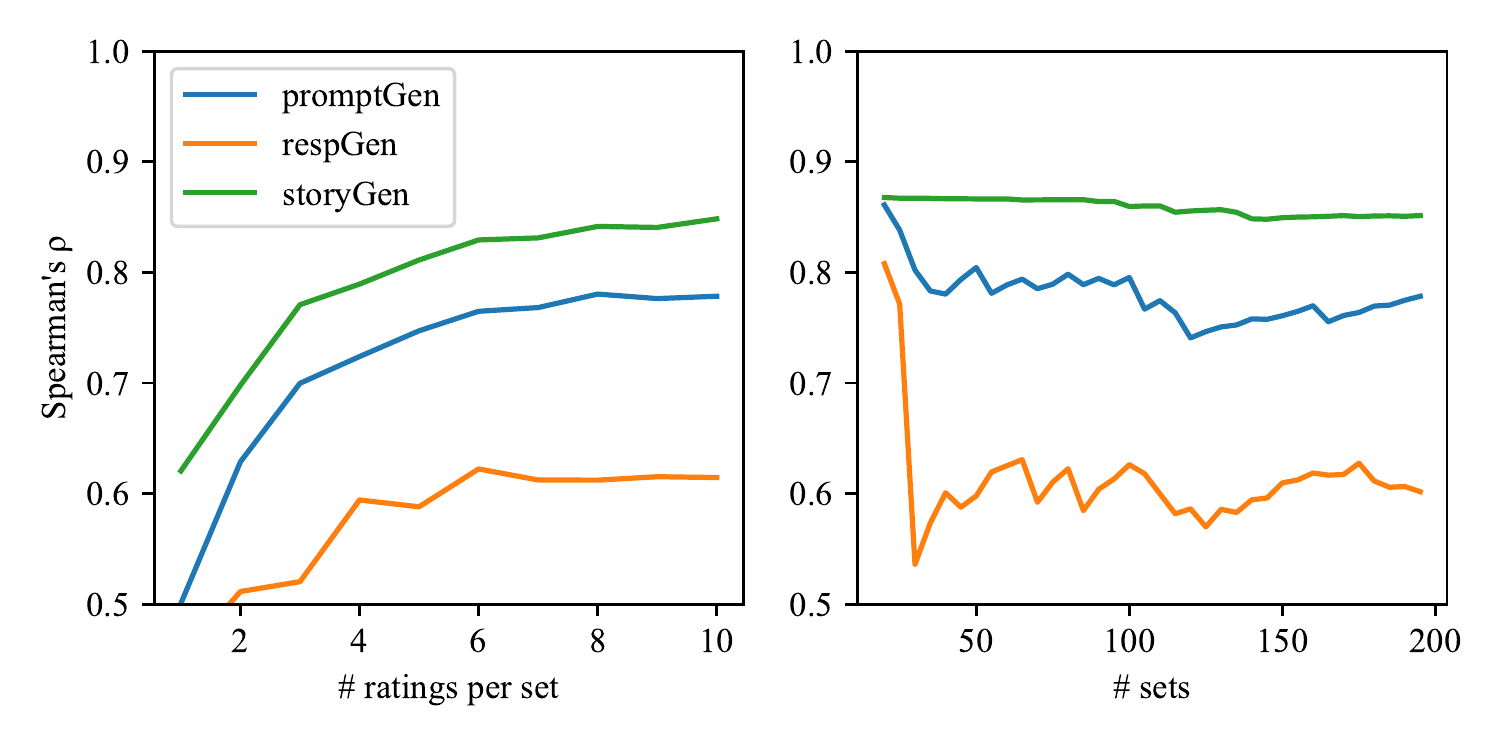}}
\caption{conTest \textit{abs\dhs} results depends on the number of ratings per set and the number of sets.}
\label{fig:stability}
\end {figure}

\paragraph{\dhs{} Stability: Picking Parameter Values}
\label{sec:exp_test_stab}
\dhs{} experiments demand expensive human labor. Thus, we need to carefully choose the number of sets and different ratings we ask per set, to get reliable results in a reasonable budget.
To this end, we conducted two series of experiments, once increasing the number of sets, and again increasing the number of ratings per sets. By observing results along those two series, we chose to use 200 sets and 10 ratings per set for all experiments - the minimal values in which results are confidently stable. Results are presented in Figure~\ref{fig:stability}.






\section{Aspects of Diversity}
\label{sec:discussion}

\label{sec:d_aspects}

In this work, we focused on the two primary aspects of diversity:   \textit{content} diversity (What to say?) and \textit{form} diversity (How to say it?). In Figure~\ref{fig:fig1}, Both sets are diverse, but \textit{Set B} is only form diverse, as all answers deliver the same massage, whereas \textit{Set A} is diverse in both form and content.

Furthermore, we can observe aspects of diversity as having a tree-like structure, where both content and form diversity can be divided to sub-aspects:
Content diversity (e.g. answering the question \nl{How are you today?}) can be expressed by using different \textit{sentiment} (\nl{I'm doing good.} vs. \nl{I'm so glad you asked! I'm really doing good.}), different \textit{relevance} (\nl{I'm fine} vs. \nl{Did you watch the game last night?}), and more. Form diversity can be divided into sub-aspects as well: \textit{syntactic} diversity (\nl{Someone took it from me.} vs. \nl{It was taken from me.}) or \textit{lexical} diversity (\nl{I feel fine.} vs. \nl{I feel very well.}). Even those sub-aspects can be further divided. For example, a sub-aspect of lexical diversity is \textit{register} diversity (\nl{How are you?} vs. \nl{Sup bro?}). 

Another observation is that different aspects are not orthogonal, that is, changing one aspect may lead to changes in other aspects. Specifically, we observe that while it is relatively easy to produce high form diversity with low content diversity (\textit{Set B} in Figure~\ref{fig:fig1}), it is almost impossible to diversify content without changing form. This observation was important during the design of conTest.



\section{Conclusions}
This work presents a framework for evaluating diversity metrics as a step toward standardized evaluation.
We limit the scope of this work to differences between \textit{form} and \textit{content} diversity, which are key towards understanding different aspects of diversity. Future work can explore other aspects of diversity,
e.g. testing \emph{sentiment} diversity, as proposed in \S\ref{sec:d_benchmark}.
We urge researchers to use this framework as a platform for developing new diversity metrics and establishing their efficiency.




\section*{Acknowledgements}
We thank Aya Meltzer-Asscher for linguistic advice, and Or Nachmias, Ben Bogin, Mor Geva, Omer Goldman and Ohad Rubin for their useful suggestions and references. This research was partially supported by The Israel Science Foundation
grant 942/16, The Yandex Initiative
for Machine Learning and the European Research
Council (ERC) under the European Union Horizons 2020 research and innovation programme
(grant ERC DELPHI 802800).

\bibliographystyle{acl_natbib}
\bibliography{ref} 

\newpage
\appendix

\section{\dhs{} Questionnaires}
\label{app:dhs_qs}
All Human scores for \dhs{} metrics were collected using Amazon Mechanical Turk (AMT) crowdsourcing platform by English native-speaking workers that were specifically qualified for this task.
Figure~\ref{app:fig:q_gen} presents the warm-up part, common for all \dhs{} questionnaires. Before asking workers to rate the diversity of each set, we first asked them to generate a response for the context themselves, to make sure they read it. To neutralize the effect of the responses' quality on the workers, we also asked the workers to rate the quality of the first response in the set, then explicitly instructed them to ignore quality when rating diversity.

Figures~\cref{app:fig:q_abs,app:fig:q_sim,app:fig:q_asp,app:fig:q_rank} present the diversity questions of absHDS, aspHDS, rnkHDS and simHDS as appeared in the AMT questionnaires.

\paragraph{Costs}
For \dhs{} metrics that require one query per response set (i.e. absHDS, rnkHDS, aspDHS), the cost for a single rating was $0.18\$$. We collected $10$ ratings per response set, and conduct each experiment with $200$ sets, hence the total cost for an experiment was $360\$$.
In the case of simHDS, the response set size was $5$, and the number of queries needed per set is $\binom{5}{2} = 10$. The cost of a single rating for this task was 0.056\$, and with the same multipliers, the total cost for an experiment was $1120\$$, three times more expensive.

\section{Data Samples}
\label{app:data_samples}

\subsection{Decoding Test (decTest)}

Tables~\cref{app:table:samples_test_1_storygen_temp,app:table:samples_test_1_storygen_topp,app:table:samples_test_1_storygen_topk,app:table:samples_test_1_respgen_temp,app:table:samples_test_1_respgen_topp,app:table:samples_test_1_respgen_topk,app:table:samples_test_1_promptgen_temp,app:table:samples_test_1_promptgen_topp,app:table:samples_test_1_promptgen_topk} present data samples from storyGen, respGen and promptGen with the neural testers of decTest, as detailed in \S\ref{sec:experiments}. Each table presents two contexts and three response sets per context. Each response set was generated with a different value of decoding parameter for the three decoding methods: softmax temperature, Nucleus sampling, and Top-k.

\subsection{Content Test (conTest)}

Tables~\cref{app:table:samples_test_2_storygen,app:table:samples_test_2_respgen,app:table:samples_test_2_promptgen} present data samples from storyGen, respGen and promptGen with the human testers of conTest, as detailed in \S\ref{sec:experiments}. Each table presents two contexts and two response sets per context - one for the \emph{low} content diversity class and one for the \emph{high} content diversity class.

\section{Additional Experiments}
\label{app:full_res}


\subsection{Decoding Test (decTest)}

\begin {figure*}
\makebox[\textwidth][c]{\includegraphics[width=0.9\textwidth]{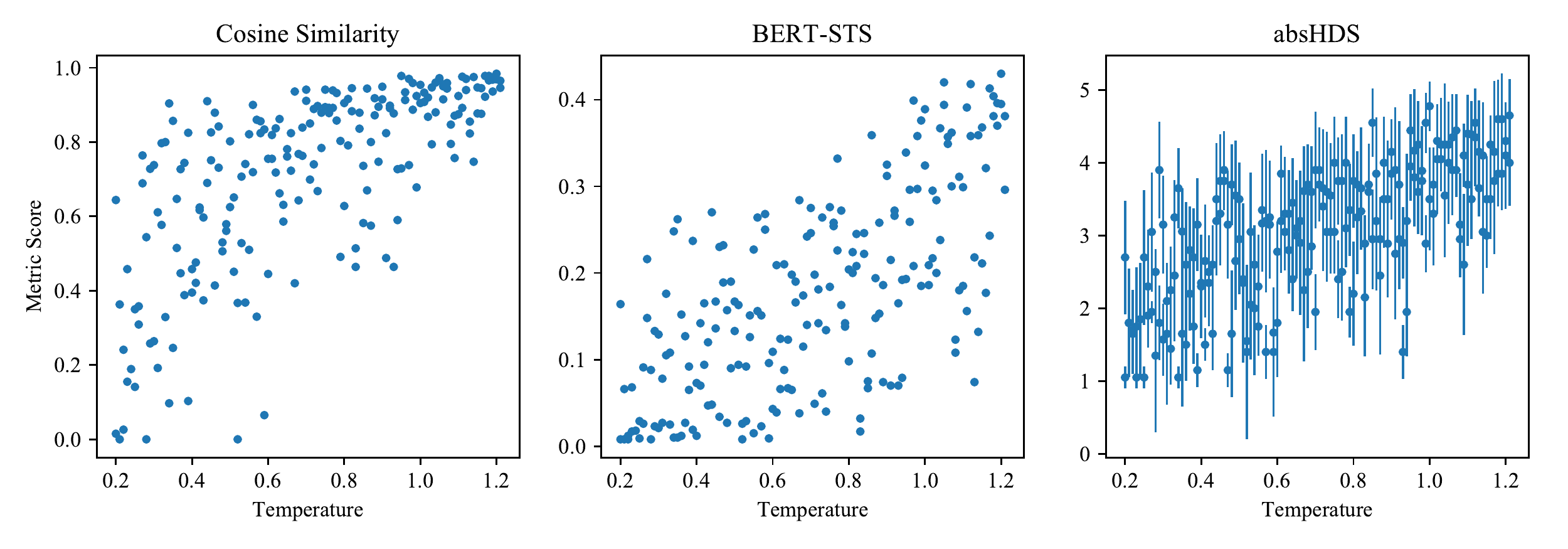}}
\caption{decTest: Scatter plot of n-gram-based (cosine similarity), neural (BERT-STS) and human (absHDS) metrics as a function of temperature for \textit{storyGen}. Each point corresponds to a single generated set. Error bars of \dhs{} represent the standard deviation over 10 annotator ratings.}
\label{app:fig:test1_storygen}
\end {figure*}

\begin {figure*}
\makebox[\textwidth][c]{\includegraphics[width=0.9\textwidth]{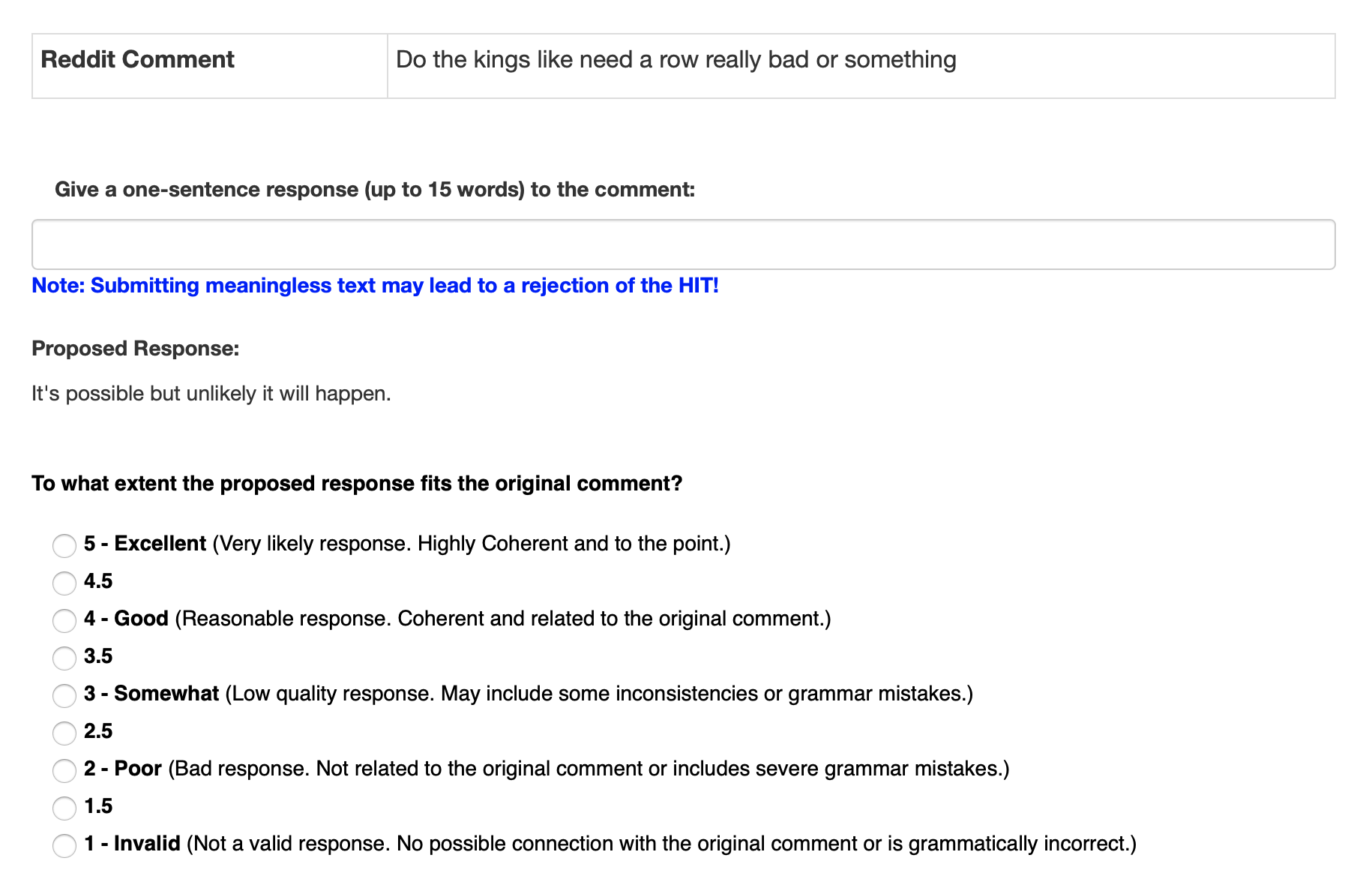}}
\caption{Warm-up part, starting each AMT \dhs{} task. It includes the context, and a single response generated by the tester. The worker is asked to generate response of hers/his own and rate the quality of the tester's response.}
\label{app:fig:q_gen}
\end {figure*}

\begin {figure*}
\makebox[\textwidth][c]{\includegraphics[width=0.9\textwidth]{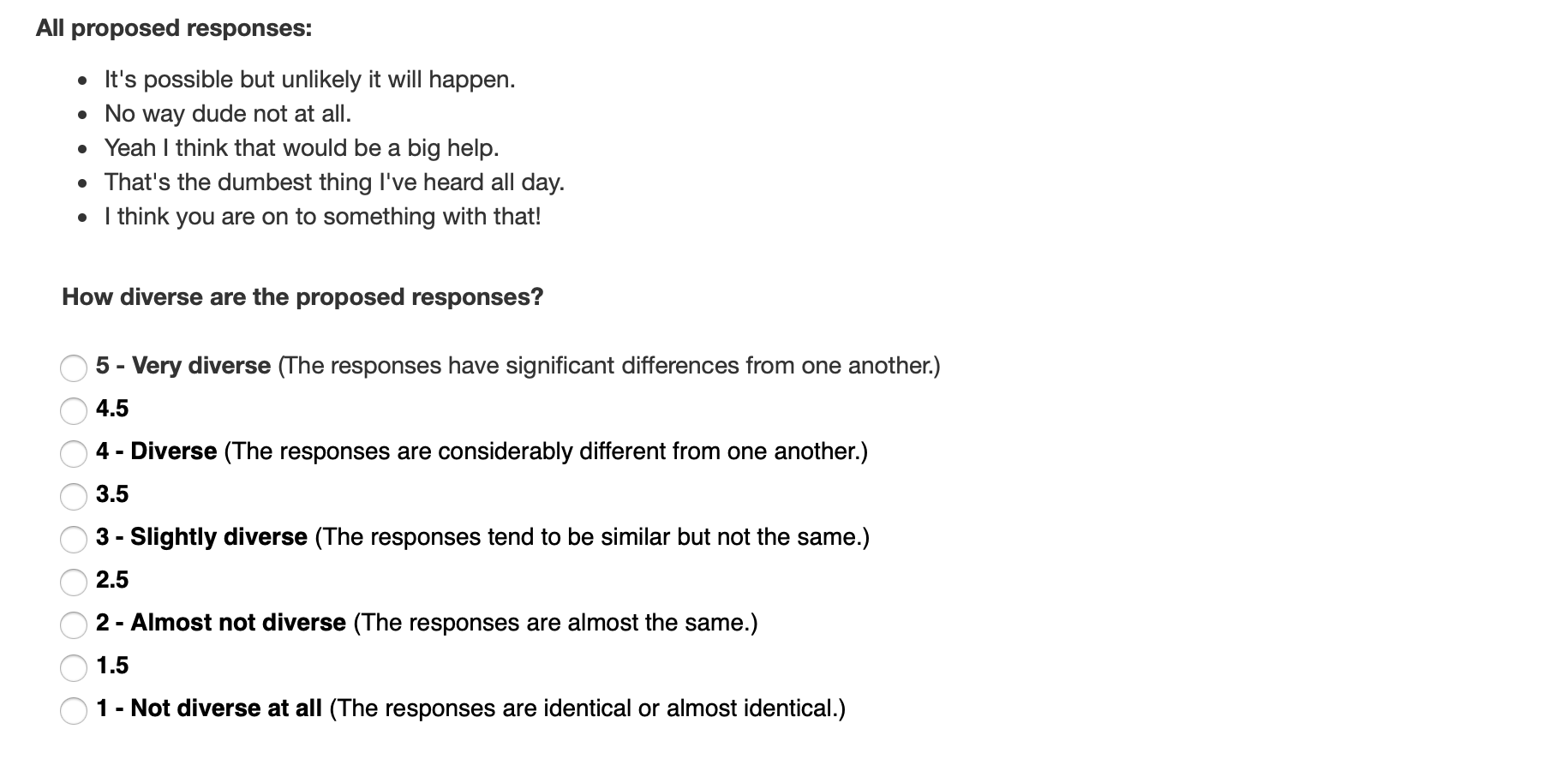}}
\caption{abs\dhs{} question along with the evaluated response set (conTest in this case).}
\label{app:fig:q_abs}
\end {figure*}

\begin {figure*}
\makebox[\textwidth][c]{\includegraphics[width=0.9\textwidth]{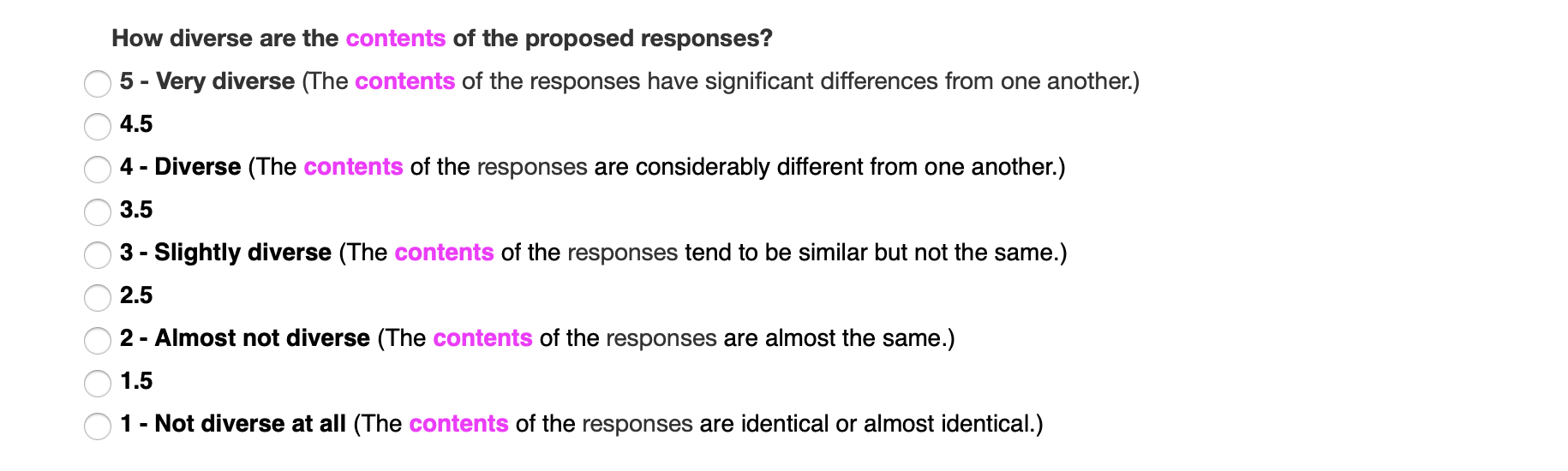}}
\caption{asp\dhs{} question (content in this case). The response set is the same as presented for abs\dhs{} question.}
\label{app:fig:q_asp}
\end {figure*}

\begin {figure*}
\makebox[\textwidth][c]{\includegraphics[width=0.9\textwidth]{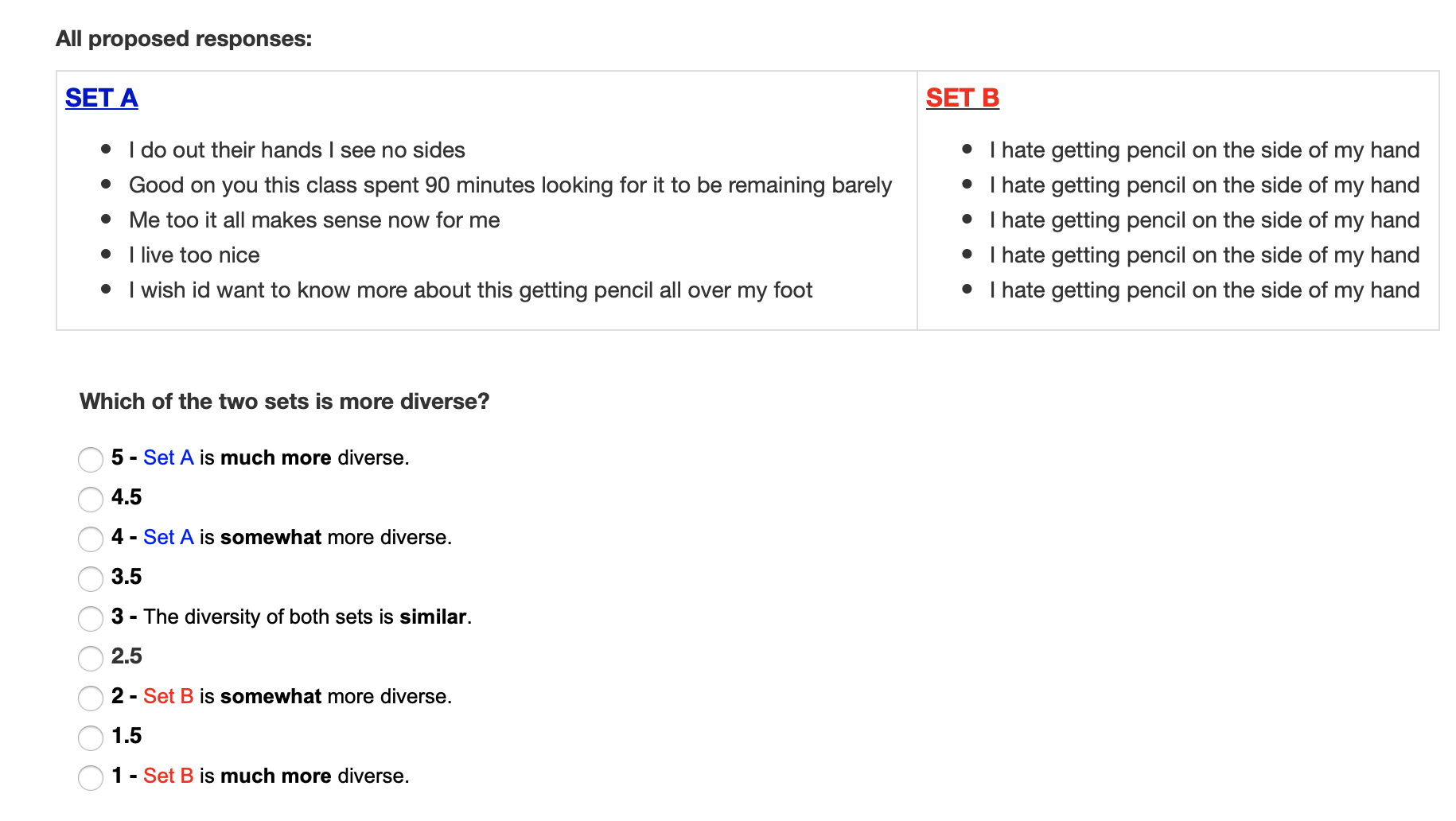}}
\caption{rnk\dhs{} question along with the two evaluated response sets.}
\label{app:fig:q_rank}
\end {figure*}

\begin {figure*}
\makebox[\textwidth][c]{\includegraphics[width=0.9\textwidth]{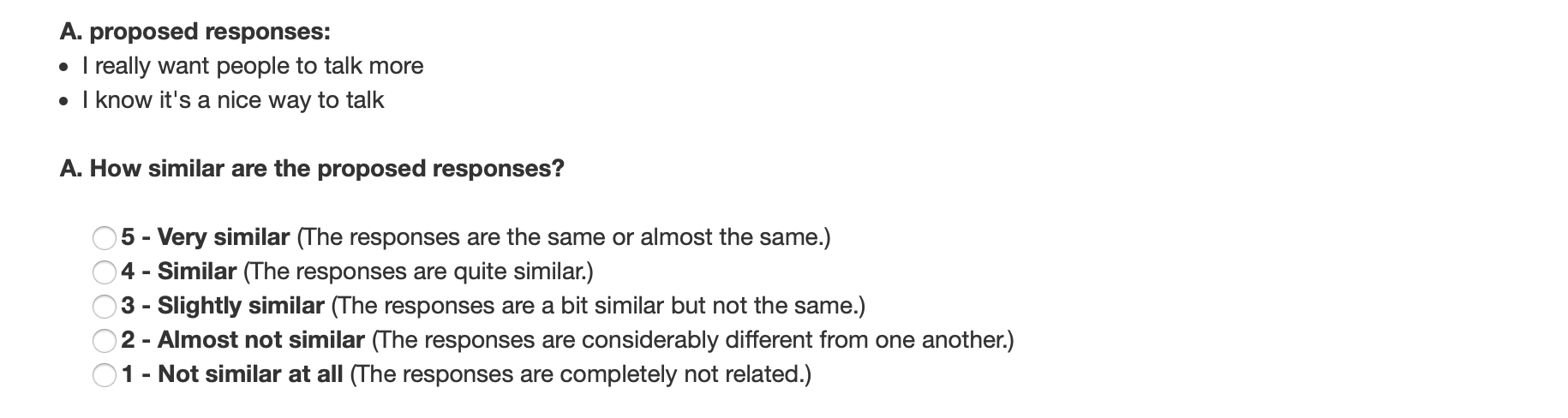}}
\caption{sim\dhs{} question along with the two evaluated responses.}
\label{app:fig:q_sim}
\end {figure*}

Comparing decTest results of storyGen to other tasks (Table~\ref{table:test1_abs}),
this task is characterised with noisier scores for all metrics (Figures \cref{app:fig:test1_storygen,figure:test1}), hence lower $\rho$ values and higher variance. A possible explanation is larger effect of $c$ on the distribution $P_{gen}(s|c)$ in this task.

Tables \cref{table:test1_d_params,app:table:test1_d_params_storygen,app:table:test1_d_params_respgen}, present decTest absolute scoring experiment using \emph{temperature}, \emph{nucleus sampling} and \emph{Top-k} decoding parameters as $d$. Top-k consistently yields lower $\rho$ compared to other decoding parameters, especially for storyGen task. This implies that Top-k 
represents diversity less reliably than other methods.

\paragraph{Ranking experiment}

To examine whether we can improve correlation by asking humans to \emph{rank} diversity, rather than providing an absolute score, we designed a ranking version of decTest. Each context is given along with two sets (5 samples each), produced with different temperature values. We sweep over temperature differences instead of the absolute temperature values. The human metric in this setting is \textit{rnk\dhs} (see \S\ref{sec:human_metrics}), and the automatic metrics are the difference between the scores each of the two sets got.

We report two measures; The first is Spearman's $\rho$ between the metric and the temperature difference.
The second is accuracy, i.e., whether the metric can predict which set has higher temperature (e.g., in automatic metrics this is whether the sign of the temperature difference and the sign of metric score difference agree).\footnote{We consider ties in the metric difference score as a miss.}

Table \ref{table:test1_rank} summarizes the ranking test results. We observe that humans are better at ranking compared to giving absolute scores (Table~\ref{table:test1_abs}), and are doing as well as automatic metrics. However, the scores of all automatic metrics also improve, making it difficult to separate between the different metrics.



\begin {table}
\centering
\resizebox {\columnwidth} {!} { %
\footnotesize

\begin{tabular}{l|cc|cc|cc} 
    \toprule
     & \multicolumn{2}{c}{\textbf{storyGen}} & \multicolumn{2}{c}{\textbf{respGen}} & \multicolumn{2}{c}{\textbf{promptGen}} \\
    & $\rho$ & acc & $\rho$ & acc & $\rho$ & acc\\
    \midrule
    distinct-n  & \textbf{0.88}  & 0.88  & 0.86  & 0.9  & \textbf{0.91}  & \textbf{0.91} \\

    cos-sim & 0.86  & 0.88  & 0.87  & \textbf{0.91}  & 0.9  & \textbf{0.91} \\
    \midrule
    BERT-STS  & 0.84  & 0.84  & 0.85  & 0.88  & 0.9  & 0.89 \\
    sent-BERT & 0.85  & 0.86  & 0.83  & 0.85  & 0.85  & 0.85 \\
    BERT-score  & \textbf{0.88}  & \textbf{0.89}  & 0.88  & 0.89  & \textbf{0.91}  & 0.9 \\

    \midrule
    rnk\dhs  & 0.87 & \textbf{0.89} & \textbf{0.89} & 0.9 & 0.89 & 0.88\\

 \bottomrule
\end{tabular}
}
\caption{\emph{decTest} ranking results: Spearman's ($\rho$) correlation between temperature differences and each metric score. Accuracy (acc) of classifying which set has the higher temperature. Standard deviation is up to $0.02$ for all automatic metrics for both Spearman's correlation and accuracy.}
\label{table:test1_rank}
\end {table}

\subsection{Metrics for Content Diversity (\mcdiv{})}

As elaborated in \S~\ref{sec:mcdiv}, \mcdiv{} is a dataset containing $6K$ $\{c,\sS_c\}$ pairs, ($2K$ for each storyGen, respGen and promptGen) collected as described in \S\ref{sec:exp_test2}. \mcdivh{} is a $3K$ subset of \mcdiv{}, in which \emph{form} diversity is neutralized, providing a difficult meta-evaluation challenge.
\mcdivh{} was sampled in a manner that causing \emph{distinct-n} metric to score zero correlation in conTest over this subset. The method of sub-sampling was meant to approximately equalize the distributions of the two classes, \emph{low} and \emph{high} content diversity, over the scores of distinct-n metric, and was performed as follows:
\begin{itemize}
    \item Sort all collected samples (from both \emph{low} and \emph{high} content diversity classes) according to their \emph{distinct-n} score.
    \item Divide the sorted samples to groups with fixed size ($40$ samples each in our case).
    \item From each such group, randomly sample \underline{the same} amount of samples for each of the two classes. For example, if a group contains $5$ \emph{low} content diversity samples and $35$ \emph{high} content diversity samples, we can sample at most $5$ samples for each class.
\end{itemize}

\paragraph{Resutls}

We applied conTest for all the collected data for each of the three NLG tasks (see Tables~\cref{app:table:mcdiv_all_results,app:table:mcdiv_hard_results}). By design, n-gram based metrics score near-zero correlation on \mcdivh{}, making \emph{high} and \emph{low} content diversity classes almost indistinguishable for those metrics, which relay on text surface level features only. Neural metrics perform strictly worse on \mcdivh{} than \mcdiv{}.
In addition, we applied conTest on 200 randomly sampled $\{c,\sS_c\}$ pairs from \mcdivh{} for respGen task (see table~\ref{app:table:mcdiv_dhs_results}). Compared to Table~\ref{table:test2}, The gap between the best performing neural metrics (sent-BERT) and abs\dhs{} was increased in favor to \dhs{} (0.04 compared to 0.1 difference in Spearman's $\rho$).




\section{Additional Reproducibility Details}
\label{app:sec:rep_details}

\paragraph{Collected data and code}
All the collected data, metric scores per samples for each of decTest and conTest, as well as code for running and visualizing the tests, are publicly available\footnote{\url{https://github.com/GuyTevet/diversity-eval}}. The collection methods are elaborated in Section~\ref{sec:experiments}.

\paragraph{Original data}
We provide additional data for the original three datasets used in Section~\ref{sec:experiments}. 

\begin{itemize}
    \item ROC Stories dataset\footnote{\url{www.cs.rochester.edu/nlp/rocstories/}} \cite{mostafazadeh-etal-2016-corpus} used for storyGen task contains $96K$/$1K$/$1K$ train/validation/test titles and five-sentence stories. We used the samples without pre-processing for both fine-tuning MASS model and generate samples for our tests.
    
    \item Reddit comment-response dataset used for respGen task contains $37M$/$1M$/$1M$ train/validation/test comment - response pairs, extracted from the social website \url{reddit.com} scraped by \url{pushshift.io} followed by the pre-process described in \cite{hashimoto2019unifying}. We used the samples without further processing for both fine-tuning MASS model and generate samples for our tests. To the best of our knowledge, this dataset is not publicly available at the moment.
    
    \item CMDC dataset\footnote{\url{www.cs.cornell.edu/~cristian/Cornell_Movie-Dialogs_Corpus.html}} \cite{Danescu-Niculescu-Mizil+Lee:11a} contains $108K$/$30K$ train/test sentence-response pairs extracted from movie scripts. We extracted the first three words from the sentences (used as contexts for the original task) to be the context of our task. We did not use this data for training since we used GPT-2 without fine-tuning for promptGen.
\end{itemize}

\paragraph{Auto-generated data}

For decTest, we used two pre-trained generative models for generating responses given the contexts:

\begin{itemize}
    \item For storyGen and respGen tasks, we used MASS\footnote{\url{github.com/microsoft/MASS}} \cite{song2019mass} (6L-1024H-8A architecture suggested by the authors), pre-traind as described in the original paper. For each task separately, we fine-tuned MASS using the training division of the dataset corresponding to the task. Fine-tuning was done using $200K$ examples over $30$ epochs, and took $23$ hours using a single \emph{TITAN Xp} GPU core. Inference with the fine-tuned model takes $65$ milliseconds on average per response set containing $10$ responses with the same GPU core.
    
    \item For promptGen task, we used Hugging-Face implementation\footnote{\url{github.com/huggingface/transformers}} of GPT-2 \emph{large} (36-layer, 1280-hidden, 20-heads, 774M parameters) \cite{radford2019language} pre-traind as described in the original paper. We used this model as-is, without fine-tuning. Inference takes $0.6$ second on average per response set containing $10$ responses with a single \emph{TITAN Xp} GPU core.
\end{itemize}

\paragraph{Tests Runtime}
Given metric scores per sample, running each of the tests with 200 samples takes less than a minute on a standard \emph{Intel i7} CPU.

\begin {table}
\centering
\resizebox {\columnwidth} {!} { %
\footnotesize

\begin{tabular}{lccc} \toprule
     & \textbf{Temperature} & \textbf{Top-p} & \textbf{Top-k} \\
    \midrule
    distinct-n  & \textbf{0.76} (0.03) & \textbf{0.69} (0.03) & 0.2 (0.06) \\
    cos-sim & 0.71 (0.04) & 0.66 (0.03) & 0.16 (0.06) \\
    \midrule
    BERT-STS & 0.64 (0.04)   & 0.58 (0.04) & 0.2 (0.07) \\
    sent-BERT  & 0.65 (0.03)   & 0.59 (0.04) & 0.17 (0.06) \\
    BERT-score  & 0.69 (0.04)   & 0.61 (0.04) & \textbf{0.23} (0.05) \\

 \bottomrule
\end{tabular}

}
\caption{decTest results for different decoding parameters: Spearman's $\rho$ (mean and standard deviation) of automatic metrics for \emph{storyGen}.}
\label{app:table:test1_d_params_storygen}
\end {table}

\begin {table}
\centering
\resizebox {\columnwidth} {!} { %
\footnotesize

\begin{tabular}{lccc} \toprule
     & \textbf{Temperature} & \textbf{Top-p} & \textbf{Top-k} \\
    \midrule
    distinct-n  & \textbf{0.89} (0.01) & \textbf{0.84} (0.02) & \textbf{0.64} (0.04) \\
    cos-sim & \textbf{0.89} (0.01) & 0.78 (0.03) & 0.62 (0.05) \\
    \midrule
    BERT-STS & 0.81 (0.02)   & 0.74 (0.03) & 0.56 (0.04) \\
    sent-BERT  & 0.80 (0.02)   & 0.63 (0.05) & 0.51 (0.04) \\
    BERT-score  & 0.87 (0.01)   & 0.77 (0.03) & 0.6 (0.05) \\

 \bottomrule
\end{tabular}
}
\caption{decTest results for different decoding parameters: Spearman's $\rho$ (mean and standard deviation) of automatic metrics for \emph{respGen}.}
\label{app:table:test1_d_params_respgen}
\end {table}

\begin {table}
\centering
\resizebox {\columnwidth} {!} { %
\footnotesize

\begin{tabular}{l|cc|cc|cc} 
    \toprule
     & \multicolumn{2}{c}{\textbf{storyGen}} & \multicolumn{2}{c}{\textbf{respGen}} & \multicolumn{2}{c}{\textbf{promptGen}} \\
    & $\rho$ & OCA & $\rho$ & OCA & $\rho$ & OCA\\
    \midrule
    distinct-n  & 0.53 & 0.74 & 0.52 & 0.74 & 0.48 & 0.75\\

    cos-sim & 0.53 & 0.74 & 0.52 & 0.74 & 0.60 & 0.77\\
    \midrule
    BERT-STS  & 0.57 & 0.74 & 0.61 & 0.78 & 0.78 & 0.89\\
    sent-BERT  & \textbf{0.75} & \textbf{0.87} & \textbf{0.68} & \textbf{0.83} & \textbf{0.8} & \textbf{0.9}\\
    BERT-score  & 0.60 & 0.77 & 0.56 & 0.78 & 0.54 & 0.74\\

 \bottomrule
\end{tabular}
}
\caption{conTest results for \mcdiv{}; Results for automatic metrics over all the samples (2K per task).}
\label{app:table:mcdiv_all_results}
\end {table}

\begin {table}
\centering
\resizebox {\columnwidth} {!} { %
\footnotesize

\begin{tabular}{l|cc|cc|cc} 
    \toprule
     & \multicolumn{2}{c}{\textbf{storyGen}} & \multicolumn{2}{c}{\textbf{respGen}} & \multicolumn{2}{c}{\textbf{promptGen}} \\
    & $\rho$ & OCA & $\rho$ & OCA & $\rho$ & OCA\\
    \midrule
    distinct-n & -0.002 & 0.49 & -0.002 & 0.49 & -0.003 & 0.49\\

    cos-sim & 0.04 & 0.53 & 0.08 & 0.55 & 0.22 & 0.60\\
    \midrule
    BERT-STS  & 0.34 & 0.64 & 0.39 & 0.68 & 0.68 & 0.83\\
    sent-BERT  & \textbf{0.63} & \textbf{0.80} & \textbf{0.53} & \textbf{0.76} & \textbf{0.73} & \textbf{0.85}\\
    BERT-score  & 0.35 & 0.66 & 0.33 & 0.65 & 0.35 & 0.65\\

 \bottomrule
\end{tabular}
}
\caption{conTest results for \mcdivh{} subset; Results for automatic metrics over all the samples (1K per task).}
\label{app:table:mcdiv_hard_results}
\end {table}

\begin {table}
\centering
\resizebox {\columnwidth} {!} { %
\footnotesize

\begin{tabular}{l|cc|cc|cc} 
    \toprule
     & \multicolumn{2}{c}{\textbf{storyGen}} & \multicolumn{2}{c}{\textbf{respGen}} & \multicolumn{2}{c}{\textbf{promptGen}} \\
    & $\rho$ & OCA & $\rho$ & OCA & $\rho$ & OCA\\
    \midrule
    distinct-n  & 0.04 & 0.57 & -0.01 & 0.46 & 0.12 & 0.56\\

    cos-sim & 0.05 & 0.54 & 0.04 & 0.54 & 0.28 & 0.62\\
    \midrule
    BERT-STS  & 0.34 & 0.68 & 0.39 & 0.69 & 0.72 & 0.9\\
    sent-BERT  & 0.68 & 0.85 & 0.6 & 0.79 & 0.75 & 0.88\\
    BERT-score  & 0.37 & 0.69 & 0.34 & 0.68 & 0.38 & 0.69\\

    \midrule
    abs\dhs  & \textbf{0.78} & \textbf{0.9} & \textbf{0.7} & \textbf{0.85} & \textbf{0.84} & \textbf{0.94}\\

 \bottomrule
\end{tabular}
}
\caption{conTest results for 200 random samples from \mcdivh{} including \dhs{}.}
\label{app:table:mcdiv_dhs_results}
\end {table}

\begin {table*}
\centering
\scriptsize

\begin{tabularx}{\linewidth}{
|>{\hsize=.25\hsize}X|
>{\hsize=.25\hsize}X|
>{\hsize=.25\hsize}X|
>{\hsize=.25\hsize}X|
}
    \hline
    \textbf{Context} & \textbf{Response set ($\tau = 0.25$)} & \textbf{Response set ($\tau = 0.8$)} & \textbf{Response set ($\tau = 1.1$)} \\
    \hline
    
    \makecell[Xt]
    {
    \textbf{Fire next door.}\\
    John woke up smelling like\\ something was burning.\\
    He went outside.\\
    He saw the fire next door.\\
    He called the authorities.} &
    
    \makecell[Xt]
    {
    {\tiny $\bullet$} It was a minor fire and they put\\ it out.\\
    {\tiny $\bullet$} It was a fire.\\
    {\tiny $\bullet$} It was a fire.\\
    {\tiny $\bullet$} It was a fire.\\
    {\tiny $\bullet$} It was a fire.\\
    {\tiny $\bullet$} It was a fire.\\
    {\tiny $\bullet$} It was a minor fire.\\
    {\tiny $\bullet$} It was a fire.\\
    {\tiny $\bullet$} It was a fire.\\
    {\tiny $\bullet$} It was a minor fire and they put\\ it out.} &
    
    \makecell[Xt]
    {
    {\tiny $\bullet$} They arrived and put out the fire. \\
    {\tiny $\bullet$} It was a fire. \\
    {\tiny $\bullet$} It was a fire. \\
    {\tiny $\bullet$} It turned out to be a fire. \\
    {\tiny $\bullet$} It was a minor fire night. \\
    {\tiny $\bullet$} They arrived and put it out. \\
    {\tiny $\bullet$} It was a scary but beautiful fire\\ next time he went in. \\
    {\tiny $\bullet$} It was a fire. \\
    {\tiny $\bullet$} It was a put out and John was able\\ to put it out. \\
    {\tiny $\bullet$} It was a scary place to live.} &
    
    \makecell[Xt]
    {
    {\tiny $\bullet$} It turned out to be a mechanic.\\
    {\tiny $\bullet$} Before the fire was put out it\\ was a fire.\\
    {\tiny $\bullet$} It was a fire.\\
    {\tiny $\bullet$} They co-worker matter how bad\\ the fire was.\\
    {\tiny $\bullet$} Several shells, the fire department\\ came just in time.\\
    {\tiny $\bullet$} They arrived and put out the fire.\\
    {\tiny $\bullet$} It turned out to be a clear, fact.\\
    {\tiny $\bullet$} It was a fire.\\
    {\tiny $\bullet$} It was a scary meal for wet clothes.\\
    {\tiny $\bullet$} It was a scary stealing from him.}\\

    \hline
    
    \makecell[lt]{
    \textbf{Charles Shaves His Chest.}\\
    Charles decides he no longer wants\\ chest hair.\\
    He shaves his chest hair with a razor.\\
    Although his skin is irritated, he hair\\ is now gone.\\
    Charles is satisfied.
    } &
    
    \makecell[lt]{
    {\tiny $\bullet$} He is glad that he finally shaved.\\
    {\tiny $\bullet$} He is glad that he finally shaved.\\
    {\tiny $\bullet$} He is very happy to have a chest\\ hair.\\
    {\tiny $\bullet$} He is glad that he finally shaved.\\
    {\tiny $\bullet$} He is glad that he finally shaved.\\
    {\tiny $\bullet$} He is glad that he finally shaved.\\
    {\tiny $\bullet$} He is glad that he finally shaved.\\
    {\tiny $\bullet$} He is glad that he finally shaved.\\
    {\tiny $\bullet$} He is glad that he finally shaved.\\
    {\tiny $\bullet$} He is glad that he finally shaved.} &
    
    \makecell[lt]{
    {\tiny $\bullet$} He is very happy to have a chest\\ hair.\\
    {\tiny $\bullet$} He is glad that he finally shaved.\\
    {\tiny $\bullet$} He is very happy that he no longer\\ has chest hair.\\
    {\tiny $\bullet$} He is glad that he finally shaved.\\
    {\tiny $\bullet$} He is glad that he finally shaved.\\
    {\tiny $\bullet$} He is glad that his chest hair is no\\ longer shaved.\\
    {\tiny $\bullet$} He is glad that he finally shaved\\
    {\tiny $\bullet$} He is glad that he finally shaved.\\
    {\tiny $\bullet$} He is glad that he finally of shaving.\\
    {\tiny $\bullet$} He is glad that he finally shashock\\ his chest hair.} &
    
    \makecell[lt]{
    {\tiny $\bullet$} He is glad that he finally peaceful\\ his voice again.\\
    {\tiny $\bullet$} He is glad that he finally shaavoided\\ his chest hair.\\
    {\tiny $\bullet$} He is glad that he finally downcome.\\
    {\tiny $\bullet$} He is ready to start better\\ cooperation his fireplace.\\
    {\tiny $\bullet$} He is glad that he finally shaved.\\
    {\tiny $\bullet$} He is glad that he no longer\\ has chest hair.\\
    {\tiny $\bullet$} He is happy to finally be shaved.\\
    {\tiny $\bullet$} He is very happy to have ster good\\ ride of his life.\\
    {\tiny $\bullet$} He is glad that he finally shaved.\\
    {\tiny $\bullet$} He is glad that he finally has chest\\ hair.}\\
    
    \hline
    
\end{tabularx}
\caption{decTest data samples for storyGen task and different temperatures.}
\label{app:table:samples_test_1_storygen_temp}
\end {table*}

\begin {table*}
\centering
\scriptsize

\begin{tabularx}{\linewidth}{
|>{\hsize=.25\hsize}X|
>{\hsize=.25\hsize}X|
>{\hsize=.25\hsize}X|
>{\hsize=.25\hsize}X|
}
    \hline
    \textbf{Context} & \textbf{Response set ($p = 0.208$)} & \textbf{Response set ($p = 0.64$)} & \textbf{Response set ($p = 1$)} \\
    \hline
    
    \makecell[lt]{
    \textbf{Really Bad Decisions.}\\
    Jake was nervous about a meeting at\\ work the next day.\\
    He decided to have a drink to relax.\\ Unfortunately Jake kept drinking.\\
    He was really hung over the next day.
    } & 
    
    \makecell[lt]{  
    {\tiny $\bullet$} He missed his meeting.\\
    {\tiny $\bullet$} He missed his meeting.\\
    {\tiny $\bullet$} He missed his meeting.\\
    {\tiny $\bullet$} He missed his meeting.\\
    {\tiny $\bullet$} He missed his meeting.\\
    {\tiny $\bullet$} He missed his meeting.\\
    {\tiny $\bullet$} He missed his meeting.\\
    {\tiny $\bullet$} He missed his meeting.\\
    {\tiny $\bullet$} He missed his meeting.\\
    {\tiny $\bullet$} He missed his meeting.\\
    } &
    
    \makecell[lt]{  
    {\tiny $\bullet$} He missed his meeting.\\
    {\tiny $\bullet$} He missed his meeting.\\
    {\tiny $\bullet$} He missed his meeting.\\
    {\tiny $\bullet$} He missed his meeting.\\
    {\tiny $\bullet$} He missed his meeting.\\
    {\tiny $\bullet$} He missed his meeting.\\
    {\tiny $\bullet$} He missed his meeting.\\
    {\tiny $\bullet$} He missed his meeting.\\
    {\tiny $\bullet$} He missed his meeting.\\
    {\tiny $\bullet$} He missed his meeting.\\
    } &
    
    \makecell[lt]{  
    {\tiny $\bullet$} He did not get to the meeting\\ anymore.\\
    {\tiny $\bullet$} He missed his meeting.\\
    {\tiny $\bullet$} He passed out and failing the\\ meeting\\
    {\tiny $\bullet$} He missed his meeting.\\
    {\tiny $\bullet$} He missed his meeting.\\
    {\tiny $\bullet$} He missed his meeting.\\
    {\tiny $\bullet$} He missed his meeting.\\
    {\tiny $\bullet$} He passed out and was kicked out\\ of the meeting.\\
    {\tiny $\bullet$} He missed his meeting.\\
    {\tiny $\bullet$} He missed his meeting.\\
    } \\
    
    \hline
    
    \makecell[lt]{
    \textbf{Family Night Food.}\\
     Tonight, my mom ordered Mexican\\ food for family night.\\
     She got it from my favorite Mexican\\ place in town.\\
     When it arrived, it was hot and\\ smelled wonderful.\\
     We devoured it with gusto.
    } & 
    
    \makecell[lt]{  
    {\tiny $\bullet$} After a few hours of take it home\\ we all enjoyed its night.\\
    {\tiny $\bullet$} After a few hours of take it home\\ we all enjoyed its night.\\
    {\tiny $\bullet$} After a few hours of take it home\\ we all enjoyed its night.\\
    {\tiny $\bullet$} After a few hours of eating\\ everyone was satisfied.\\
    {\tiny $\bullet$} After a few hours of take it home\\ we all enjoyed its night.\\
    {\tiny $\bullet$} After a few hours of eating \\ everyone was satisfied.\\
    {\tiny $\bullet$} After a few hours of take it home\\ we all enjoyed its night.\\
    {\tiny $\bullet$} After a few hours of take it home\\ we all enjoyed its night.\\
    {\tiny $\bullet$} After a few hours of take it home\\ we all enjoyed its night.\\
    {\tiny $\bullet$} After a few hours of eating\\ everyone was satisfied.\\
    } &
    
    \makecell[lt]{  
    {\tiny $\bullet$} After dinner, we all went home to\\ cook Mexican food.\\
    {\tiny $\bullet$} After a few hours of cooking she\\ was tired and ready to eat.\\
    {\tiny $\bullet$} After dinner, I always put got ready\\ for Christmas.\\
    {\tiny $\bullet$} After dinner, I helped her do the\\ dishes.\\
    {\tiny $\bullet$} After a few hours of dinner, the\\ food was amazing.\\
    {\tiny $\bullet$} After a few hours of take it home\\ we all enjoyed margaria.\\
    {\tiny $\bullet$} After dinner, I was ready to take\\ on work the next day of\\
    {\tiny $\bullet$} After dinner, I was sad to say\\ goodbye to her\\
    {\tiny $\bullet$} After a few hours of take it home\\ we all enjoyed one bite.\\
    {\tiny $\bullet$} After a few hours of eating\\ everyone was satisfied. \\
    } &
    
    \makecell[lt]{  
    {\tiny $\bullet$} Even though my stomach was gone,\\ I was sad it was finally pockets\\
    {\tiny $\bullet$} After dinner, I alone.\\
    {\tiny $\bullet$} All in all the family while my mom\\ finished the food.\\
    {\tiny $\bullet$} After a few hours of dancing, she\\ pianed.\\
    {\tiny $\bullet$} Afterwards I'd never finish single\\ night a week.\\
    {\tiny $\bullet$} Afterwards we all went to sleep\\ in my woods.\\
    {\tiny $\bullet$} Afterwards I helped her do much\\ better than my wife.\\
    {\tiny $\bullet$} Afterwards my mom helped me\\ finish my dinner.\\
    {\tiny $\bullet$} After a few hours we all enjoyed\\ storm blood from the Italy.\\
    {\tiny $\bullet$} After dinner, we all enjoyed some\\ good food together.\\
    } \\
    
    \hline

\end{tabularx}
\caption{decTest data samples for storyGen task and different $p$ values (nucleus sampling).}
\label{app:table:samples_test_1_storygen_topp}
\end {table*}

\begin {table*}
\centering
\scriptsize

\begin{tabularx}{\linewidth}{
|>{\hsize=.25\hsize}X|
>{\hsize=.25\hsize}X|
>{\hsize=.25\hsize}X|
>{\hsize=.25\hsize}X|
}

    \hline
    \textbf{Context} & \textbf{Response set ($k = 3$)} & \textbf{Response set ($k = 32$)} & \textbf{Response set ($k = 318$)} \\
    \hline
    
    \makecell[lt]{
     \textbf{Loud Noise.} \\
     Jane was trying to take a nap. \\
     She heard a loud bang in the kitchen. \\
     It woke her up. \\
     A dish had fallen off the counter.
    } & 
    
    \makecell[lt]{  
    {\tiny $\bullet$} Jane pulled over and started to\\ clean herself.\\
    {\tiny $\bullet$} Jane was horrified and dropped her\\ favorite food.\\
    {\tiny $\bullet$} Jane was able to finish her car and\\ take a nap.\\
    {\tiny $\bullet$} Jane was able to finish her nap .\\
    {\tiny $\bullet$} Jane was able to finish her nap .\\
    {\tiny $\bullet$} Jane stopped at the store to buy\\ a new one.\\
    {\tiny $\bullet$} Jane was able to grab a nap\\ afterwards.\\
    {\tiny $\bullet$} Jane was unable to finish her nap\\ since it was lost.\\
    {\tiny $\bullet$} Jane pulled over and started to\\ clean up.\\
    {\tiny $\bullet$} Jane was able to finish her nap .\\
    } &
    
    \makecell[lt]{  
    {\tiny $\bullet$} Jane was able to catch a car using\\ the seat.\\
    {\tiny $\bullet$} Jane stopped at the store to buy\\ a new book.\\
    {\tiny $\bullet$} Jane was sad her cat dropped out\\ of the kitchen.\\
    {\tiny $\bullet$} Jane screamed.\\
    {\tiny $\bullet$} Jane was horrified to find her car\\ broken down on the floor.\\
    {\tiny $\bullet$} Jane was horrified and dropped her\\ pay phone.\\
    {\tiny $\bullet$} Jane was easily able to grab a nap.\\
    {\tiny $\bullet$} Jane pulled over and started to cry.\\
    {\tiny $\bullet$} Jane pulled over and started to cry.\\
    {\tiny $\bullet$} Jane stopped at the store to buy a\\ new dish from the store.\\
    } &
    
    \makecell[lt]{  
    {\tiny $\bullet$} Jane comes, noticed a lot of food\\ left under it.\\
    {\tiny $\bullet$} Jane was horrified and dropped\\ her book.\\
    {\tiny $\bullet$} Jane remembered to take a nap.\\
    {\tiny $\bullet$} Jane was since she took a nap while\\ she waited for the refun\\
    {\tiny $\bullet$} Jane knew she had no time to finish\\ her book.\\
    {\tiny $\bullet$} Jane was glad.\\
    {\tiny $\bullet$} Jane was annoyed and began to cry.\\
    {\tiny $\bullet$} Jane stopped at the store to buy a\\ new one.\\
    {\tiny $\bullet$} Jane wanted to have her car back.\\
    {\tiny $\bullet$} Jane was monthed.\\
    } \\
    
    \hline
    
    \makecell[lt]{
     \textbf{Headache.}\\
     Kate was wearing big over the ear\\ headphones.\\
     But they were tight and squeezing\\ her head.\\
     She tried to adjust them to relieve\\ the tension.\\
     But nothing really worked.
    } & 
    
    \makecell[lt]{  
    {\tiny $\bullet$} Kate decided to go to the store\\ and buy some ear phones.\\
    {\tiny $\bullet$} She decided to go to the store\\ and buy some headphones instead.\\
    {\tiny $\bullet$} She decided to go to the store\\ and buy some ear phones.\\
    {\tiny $\bullet$} She decided to go buy a pair of\\ headphones instead.\\
    {\tiny $\bullet$} She decided to go to the store and\\ buy some headphones instead.\\
    {\tiny $\bullet$} She decided to go to the store and\\ buy some headphones instead.\\
    {\tiny $\bullet$} She decided to go to the doctor\\ and have some rest.\\
    {\tiny $\bullet$} Kate decided to go to the store\\ and buy some headphones instead.\\
    {\tiny $\bullet$} She decided to go to the store and\\ buy some ear phones.\\
    {\tiny $\bullet$} She decided to go buy some ear\\ plugs.\\
    } &
    
    \makecell[lt]{  
    {\tiny $\bullet$} She decided to go buy a hat so she\\ could enjoy the the long lin\\
    {\tiny $\bullet$} Kate decided to go to the store to\\ buy some candy since she was ti\\
    {\tiny $\bullet$} Kate decided to go outside to rest.\\
    {\tiny $\bullet$} Kate decided to go to the store and\\ buy some ran phones.\\
    {\tiny $\bullet$} Kate decided to go to the store and\\ buy some headphones instead.\\
    {\tiny $\bullet$} Kate decided to go back to her old\\ dishes.\\
    {\tiny $\bullet$} She decided to go buy a big pair\\ of headphones instead.\\
    {\tiny $\bullet$} Kate decided to go to the store and\\ wearing some headphones.\\
    {\tiny $\bullet$} She was forced to go to the store\\ to buy some cash.\\
    {\tiny $\bullet$} She decided to go to the store and\\ buy some headphones instead.\\
    } &
    
    \makecell[lt]{  
    {\tiny $\bullet$} She decided to go to a ess instead.\\
    {\tiny $\bullet$} Kate decided to go to the store and\\ buy some fake headphones.\\
    {\tiny $\bullet$} She decided to go to the store and\\ buy some ear phones.\\
    {\tiny $\bullet$} She decided to go to the store and\\ buy some headphones instead.\\
    {\tiny $\bullet$} She decided to go to the store and\\ buy some headphones instead.\\
    {\tiny $\bullet$} She decided to go buy some ear\\ phones.\\
    {\tiny $\bullet$} Kate decided to go to the gym and\\ use some saw no more watching T\\
    {\tiny $\bullet$} Kate decided to go to the store and\\ buy some cute phones.\\
    {\tiny $\bullet$} She decided to go buy a fake\\ headphones instead.\\
    {\tiny $\bullet$} She decided to go to the store and\\ buy some concert.\\
    } \\
    \hline
    
\end{tabularx}
\caption{decTest data samples for storyGen task and different $k$ values (Top-k).}
\label{app:table:samples_test_1_storygen_topk}
\end {table*}

\begin {table*}
\centering
\scriptsize

\begin{tabularx}{\linewidth}{
|>{\hsize=.25\hsize}X|
>{\hsize=.25\hsize}X|
>{\hsize=.25\hsize}X|
>{\hsize=.25\hsize}X|
}
    \hline
    \textbf{Context} & \textbf{Response set ($\tau = 0.25$)} & \textbf{Response set ($\tau = 0.8$)} & \textbf{Response set ($\tau = 1.1$)} \\
    \hline
    
    \makecell[lt]{
    it is doing okay i just want people to\\ talk more
    } &
    \makecell[lt]{
    {\tiny $\bullet$} i think you mean people\\
    {\tiny $\bullet$} i think its working now\\
    {\tiny $\bullet$} i just want people to talk more\\
    {\tiny $\bullet$} i think i just want people to talk\\ more\\
    {\tiny $\bullet$} i dont think you understand what\\ youre saying\\
    {\tiny $\bullet$} i like to think so\\
    {\tiny $\bullet$} i know that feel\\
    {\tiny $\bullet$} i dont know what to say\\
    {\tiny $\bullet$} i think its pretty cool\\
    {\tiny $\bullet$} i think its working now\\
    } &
    
    \makecell[lt]{
    {\tiny $\bullet$} i cant believe you didnt think\\ this through i actually havent had\\
    {\tiny $\bullet$} may i ask why\\
    {\tiny $\bullet$} i just hope its not going to happen\\
    {\tiny $\bullet$} its alright dont even want to talk\\
    {\tiny $\bullet$} what did it say i was using\\ windows 95 anyway xd\\
    {\tiny $\bullet$} thanks for reading the thread\\
    {\tiny $\bullet$} i understand and hope they dont\\ mind that\\
    {\tiny $\bullet$} want to talk more\\
    {\tiny $\bullet$} its doing ok body language great\\ voice acting and music\\
    {\tiny $\bullet$} all good i know your pain\\
    } &
    
    \makecell[lt]{
    {\tiny $\bullet$} im just hawaii everyone wants\\ to be friends lt go talk to people from\\
    {\tiny $\bullet$} its pretty smart we didnt run\\ a year ago and have turned back but\\
    {\tiny $\bullet$} id lick your teeth but tha\\t sucks smart people\\
    {\tiny $\bullet$} it really didnt stop i think\\ you need to go back to normal now\\
    {\tiny $\bullet$} sounds like a plan to me\\
    {\tiny $\bullet$} we we can try forever practice\\ beer\\
    {\tiny $\bullet$} okay but another two more 200\\ radio views im going to ore my\\ comment of\\
    {\tiny $\bullet$} team building looking bro\\
    {\tiny $\bullet$} yep they said that things have\\ changed\\
    {\tiny $\bullet$} yeah we thank you random stranger\\
    }\\
    
    \hline
    
    \makecell[lt]{
    girls who say no  not like it matters
    } &
    \makecell[lt]{
    {\tiny $\bullet$} i dont like it at all\\
    {\tiny $\bullet$} i dont like it either\\
    {\tiny $\bullet$} i dont like it either\\
    {\tiny $\bullet$} i think you mean girls who say\\ no not like it matters\\
    {\tiny $\bullet$} i think that is the most\\ accurate description of this thread\\
    {\tiny $\bullet$} i like this one\\
    {\tiny $\bullet$} girls who say no dont like it\\ matter\\
    {\tiny $\bullet$} i dont like it either\\
    {\tiny $\bullet$} i know a guy who says no and\\ he doesnt like it\\
    {\tiny $\bullet$} girls who say no dont like it\\ matter\\
    } &
    
    \makecell[lt]{
    {\tiny $\bullet$} wait so there are girls who say no\\ like it matters\\
    {\tiny $\bullet$} i have a friend who doesnt like it\\ but i always say no\\
    {\tiny $\bullet$} and you dont like to play with\\ girls\\
    {\tiny $\bullet$} i say no i dont want it to\\
    {\tiny $\bullet$} girls who say no dont like it care\\
    {\tiny $\bullet$} i see what you did there\\
    {\tiny $\bullet$} girls who say no need to make eye\\ contact with girls\\
    {\tiny $\bullet$} girls who say yes dont like it does\\
    {\tiny $\bullet$} i really dont care if its not gay or\\ straight out programs or not\\
    {\tiny $\bullet$} ugh that game was just awful troll\\
    } &
    
    \makecell[lt]{
    {\tiny $\bullet$} how about mine like sales figure\\ instead of higher\\
    {\tiny $\bullet$} definitely not as stupid as that\\
    {\tiny $\bullet$} agreed is that true\\
    {\tiny $\bullet$} refer to my highest iq you and me\\
    {\tiny $\bullet$} less keep it up\\
    {\tiny $\bullet$} dude ive never seen such an\\ complaining member with a\\ legitimate thought\\
    {\tiny $\bullet$} not that theres anything wrong\\ with that or are more wise than\\ telling want\\
    {\tiny $\bullet$} girls make videos youve never met\\
    {\tiny $\bullet$} gtthe girls who say no or a one\\ time thing people thats\\
    {\tiny $\bullet$} which is weird since it is a girl\\ i know so sad\\
    }\\
    
    \hline
    
\end{tabularx}
\caption{decTest data samples for respGen task and different temperatures.}
\label{app:table:samples_test_1_respgen_temp}
\end {table*}

\begin {table*}
\centering
\scriptsize

\begin{tabularx}{\linewidth}{
|>{\hsize=.25\hsize}X|
>{\hsize=.25\hsize}X|
>{\hsize=.25\hsize}X|
>{\hsize=.25\hsize}X|
}
    \hline
    \textbf{Context} & \textbf{Response set ($p = 0.208$)} & \textbf{Response set ($p = 0.64$)} & \textbf{Response set ($p = 1$)} \\
    \hline
    
    \makecell[lt]{
    do you know if he still lives in new\\ orleans
    } &
    \makecell[lt]{
    {\tiny $\bullet$} i dont think so\\
    {\tiny $\bullet$} i dont think so\\
    {\tiny $\bullet$} i dont think so\\
    {\tiny $\bullet$} no i dont think so\\
    {\tiny $\bullet$} no idea i dont know\\
    {\tiny $\bullet$} i dont think so\\
    {\tiny $\bullet$} no i dont think so\\
    {\tiny $\bullet$} no idea i just saw him live in\\ new orleans\\
    {\tiny $\bullet$} i dont think so\\
    {\tiny $\bullet$} no idea i just saw him live in\\ new orleans\\
    } &
    
    \makecell[lt]{
    {\tiny $\bullet$} he is in nyc\\
    {\tiny $\bullet$} he does i hope he does\\
    {\tiny $\bullet$} i dont think so\\
    {\tiny $\bullet$} he lives in new orleans\\
    {\tiny $\bullet$} yes he does he is in florida\\
    {\tiny $\bullet$} im guessing not that i know of\\
    {\tiny $\bullet$} hes a small town i dont know what\\ happened there\\
    {\tiny $\bullet$} nope still in new orleans\\
    {\tiny $\bullet$} i do not i dont know if he still lives\\ in new orleans\\
    {\tiny $\bullet$} i dont think so\\
    } &
    
    \makecell[lt]{
    {\tiny $\bullet$} aw screw you guys the gimme a\\ shower and ill send you my money\\
    {\tiny $\bullet$} he does im so confused by how this\\ is relevant\\
    {\tiny $\bullet$} they spent all that much in new \\orleans up in first place\\
    {\tiny $\bullet$} still living in new orleans according\\ to most of them it jesus mustve worth\\
    {\tiny $\bullet$} anyone else want to wait for him \\to go back home\\
    {\tiny $\bullet$} how many tickets would you say the \\willing said if he warn you that hes\\
    {\tiny $\bullet$} im guessing he lives there as well\\
    {\tiny $\bullet$} yep rick albeit not\\
    {\tiny $\bullet$} he lives in ny\\
    {\tiny $\bullet$} no i dont get it any more guess i\\ may canadian\\
    }\\
    
    \hline
    
    \makecell[lt]{
    its that hanging thing in the back\\ of your mouth
    } &
    \makecell[lt]{
    {\tiny $\bullet$} its a tattoo of a hand\\
    {\tiny $\bullet$} its a good thing its not\\
    {\tiny $\bullet$} i think its a banana\\
    {\tiny $\bullet$} i was thinking the same thing\\
    {\tiny $\bullet$} its a very nice thing to do\\
    {\tiny $\bullet$} its a hole in the bottom of my\\ mouth\\
    {\tiny $\bullet$} i think its a good thing\\
    {\tiny $\bullet$} its a pretty cool thing to do\\
    {\tiny $\bullet$} thats what i thought\\
    {\tiny $\bullet$} i think its a butt\\
    } &
    
    \makecell[lt]{
    {\tiny $\bullet$} well it is kind of like a mac\\
    {\tiny $\bullet$} that would be a good idea\\
    {\tiny $\bullet$} its my cake day\\
    {\tiny $\bullet$} you can see it in his mouth\\
    {\tiny $\bullet$} its the junk that was pooping in\\
    {\tiny $\bullet$} i was thinking the same thing\\
    {\tiny $\bullet$} i always see that\\
    {\tiny $\bullet$} i didnt know that was possible\\
    {\tiny $\bullet$} this is my favorite comment of\\ the day\\
    {\tiny $\bullet$} thats not a sticker\\
    } &
    
    \makecell[lt]{
    {\tiny $\bullet$} i think you just flash\\
    {\tiny $\bullet$} it is and feels good\\
    {\tiny $\bullet$} my ex and i reach our arms cheating\\ into the door for almost a minute\\
    {\tiny $\bullet$} my mind went to side with that\\
    {\tiny $\bullet$} yep terrible for my teeth\\
    {\tiny $\bullet$} exactly what i was thinking\\
    {\tiny $\bullet$} why do i look so contest\\
    {\tiny $\bullet$} and where else are you going\\
    {\tiny $\bullet$} just dont wear a helmet\\
    {\tiny $\bullet$} and it looks like something some\\ one would give to get in on\\
    }\\
    
    \hline
    
\end{tabularx}
\caption{decTest data samples for respGen task and different $p$ values (nucleus sampling).}
\label{app:table:samples_test_1_respgen_topp}
\end {table*}

\begin {table*}
\centering
\scriptsize

\begin{tabularx}{\linewidth}{
|>{\hsize=.25\hsize}X|
>{\hsize=.25\hsize}X|
>{\hsize=.25\hsize}X|
>{\hsize=.25\hsize}X|
}
    \hline
    \textbf{Context} & \textbf{Response set ($k = 3$)} & \textbf{Response set ($k = 32$)} & \textbf{Response set ($k = 318$)} \\
    \hline
    
    \makecell[lt]{
    watching curry play in his prime is \\truly a privilege
    } &
    \makecell[lt]{
    {\tiny $\bullet$} i know i just dont want him to \\play for us\\
    {\tiny $\bullet$} he has to be a good center for that\\
    {\tiny $\bullet$} he is a great center of football\\ in his prime\\
    {\tiny $\bullet$} hes been playing in his prime \\for a long time\\
    {\tiny $\bullet$} he was a great back in the day\\
    {\tiny $\bullet$} hes been playing for a while now\\
    {\tiny $\bullet$} i dont know about that he was \\pretty damn good at that\\
    {\tiny $\bullet$} i dont think he was ever in his\\ prime\\
    {\tiny $\bullet$} i dont think he is a prime minister\\
    {\tiny $\bullet$} i dont know why but i think he is\\ a very good player and\\
    } &
    
    \makecell[lt]{
    {\tiny $\bullet$} and his career as well\\
    {\tiny $\bullet$} agreed the way hes playing is \\awesome\\
    {\tiny $\bullet$} it has to be\\
    {\tiny $\bullet$} this is just called a job\\
    {\tiny $\bullet$} and then being on the field for the \\first time\\
    {\tiny $\bullet$} i dont see him doing that often \\enough\\
    {\tiny $\bullet$} he just likes to party in the kitchen\\
    {\tiny $\bullet$} at this point hes going to be a great \\star for the rest of the\\
    {\tiny $\bullet$} only if he pays well\\
    {\tiny $\bullet$} the only thing that can make that\\ kind of difference is how much\\ time you\\
    } &
    
    \makecell[lt]{
    {\tiny $\bullet$} yeah my feeling i mean we dont \\like it but it happens all the\\
    {\tiny $\bullet$} you are one for real\\
    {\tiny $\bullet$} they still have a rule saying \\they might not be injured yet\\
    {\tiny $\bullet$} it really is a necessary thing to do\\
    {\tiny $\bullet$} finally some reason to continue \\watching him at some point\\
    {\tiny $\bullet$} yet that would be epic\\
    {\tiny $\bullet$} not to mention eating curry dinner\\ is a privilege\\
    {\tiny $\bullet$} i just dont want to turn over for this\\ goal like he does in\\
    {\tiny $\bullet$} gt playing in his prime is truly a\\ privilege ftfy\\
    {\tiny $\bullet$} so is saying he is in high school\\
    }\\
    
    \hline
    
    \makecell[lt]{
im going this evening when she\\ usually works 
    } &
    \makecell[lt]{
    {\tiny $\bullet$} i think you accidentally a word\\
    {\tiny $\bullet$} you are a good man\\
    {\tiny $\bullet$} i hope she works\\
    {\tiny $\bullet$} im going to the same time as you\\ when she usually works\\
    {\tiny $\bullet$} i am so sorry to hear that\\
    {\tiny $\bullet$} i hope she works for you\\
    {\tiny $\bullet$} i am so jealous of your work i \\am so jealous\\
    {\tiny $\bullet$} i hope you have fun\\
    {\tiny $\bullet$} i hope you get a job at a local bar\\
    {\tiny $\bullet$} i hope she works for you\\
    } &
    
    \makecell[lt]{
    {\tiny $\bullet$} i see what you did there\\
    {\tiny $\bullet$} where in the world are you going to\\ put your socks on\\
    {\tiny $\bullet$} she usually does but she has to keep\\ up to date with her in the\\
    {\tiny $\bullet$} awesome what do you want for it\\
    {\tiny $\bullet$} how does a women have a \\relationship with someone\\
    {\tiny $\bullet$} did you get the job\\
    {\tiny $\bullet$} thats where she goes\\
    {\tiny $\bullet$} i know im also going this friday\\
    {\tiny $\bullet$} thats a great idea\\
    {\tiny $\bullet$} well make sure youre there too good\\ luck\\
    } &
    
    \makecell[lt]{
    {\tiny $\bullet$} thats what you like to hear\\
    {\tiny $\bullet$} go see her later this evening\\
    {\tiny $\bullet$} i read that as she usually fails\\
    {\tiny $\bullet$} okay ill see you there brah\\
    {\tiny $\bullet$} i always thought that it was for the\\ sake of having the girl play as\\
    {\tiny $\bullet$} that can be expected here\\
    {\tiny $\bullet$} thats it im down now\\
    {\tiny $\bullet$} i dont see why it would matter if\\ shes married\\
    {\tiny $\bullet$} i will be the ex gf\\
    {\tiny $\bullet$} she still does make her phone \\calls the whole office\\
    }\\
    
    \hline
    
\end{tabularx}
\caption{decTest data samples for respGen task and different $k$ values (Top-k).}
\label{app:table:samples_test_1_respgen_topk}
\end {table*}

\begin {table*}
\centering
\scriptsize

\begin{tabularx}{\linewidth}{
|>{\hsize=.33\hsize}X|
>{\hsize=.33\hsize}X|
>{\hsize=.34\hsize}X|
}
    \hline
    \textbf{Response set ($\tau = 0.25$)} & \textbf{Response set ($\tau = 0.8$)} & \textbf{Response set ($\tau = 1.1$)} \\
    \hline

    \makecell[lt]{
    {\tiny $\bullet$} \textbf{Not the hacking}.  The hacking is the fact that\\ the DNC was hacked.  !\\
    {\tiny $\bullet$} \textbf{Not the hacking}.  The hacking is the real \\problem.  The hacking is the!\\
    {\tiny $\bullet$} \textbf{Not the hacking} of the DNC, but the leaks of \\the emails of the Democratic National\\ Committee.!\\
    {\tiny $\bullet$} \textbf{Not the hacking}, but the way it was done.  The\\ FBI's investigation into the!\\
    {\tiny $\bullet$} \textbf{Not the hacking} of the DNC, but the hacking \\of the emails of the Democratic National\\ Committee.!\\
    {\tiny $\bullet$} \textbf{Not the hacking} of the DNC, but the leaking \\of the emails.  The DNC's!\\
    {\tiny $\bullet$} \textbf{Not the hacking} of the DNC.  The hacking of\\ the DNC was a "false flag!\\
    {\tiny $\bullet$} \textbf{Not the hacking} of the DNC. But the hacking \\of the RNC.  The DNC hack!\\
    {\tiny $\bullet$} \textbf{Not the hacking}.  The hacking is the problem.\\  The hacking is the problem!\\
    {\tiny $\bullet$} \textbf{Not the hacking} of the DNC, but the leaking of\\ the emails. The DNC was hacked,!\\
    } &
    
    \makecell[lt]{
     {\tiny $\bullet$} \textbf{Not the hacking} after all?  I'm sure the \\nation-states that are involved in!\\
    {\tiny $\bullet$} \textbf{Not the hacking} that happened on the internal\\ networks of the Energy Department. In fact,\\ according to!\\
    {\tiny $\bullet$} \textbf{Not the hacking} of the American public but \\rather the fraudulent Heisenberg principle that\\ seemed to be!\\
    {\tiny $\bullet$} \textbf{Not the hacking} that took place in the DNC \\last year or the release of hacked emails during the!\\
    {\tiny $\bullet$} \textbf{Not the hacking} futurists  Cardboard inventor \\and self-described tinkerer Dennis!\\
    {\tiny $\bullet$} \textbf{Not the hacking} alone. In the first half of \\the report, the hackers tried to create fake!\\
    {\tiny $\bullet$} \textbf{Not the hacking}.  The hacking is the NSA's new \\SHIELD technology. It is!\\
    {\tiny $\bullet$} \textbf{Not the hacking} and hacking and hacking of the\\ world government. I know this man is a man!\\
    {\tiny $\bullet$} \textbf{Not the hacking} aspect, but the pressure exerted\\ by the Trumpistas. But also the Russia angle!\\
    {\tiny $\bullet$} \textbf{Not the hacking}, but the willingness."  The \\evidence of interest in this case comes in!\\
    } &
    
    \makecell[lt]{
    {\tiny $\bullet$} \textbf{Not the hacking} experience of a CIA VRO \\crunch nine months ago—JumpStart for 2016 jumps!\\
    {\tiny $\bullet$} \textbf{Not the hacking}, David.) The directory was\\ flagged in a document it created in late last year!\\
    {\tiny $\bullet$} \textbf{Not the hacking} of Democratic Party systems\\ - said the Russian team's activity represented \\"just the beginning!\\
    {\tiny $\bullet$} \textbf{Not the hacking}, of course -- which these\\ sources sounded more concerned about than \\being attacked 140 times!\\
    {\tiny $\bullet$} \textbf{Not the hacking} story is over. But yet \\there's another reason not to rush out such \\statements!\\
    {\tiny $\bullet$} \textbf{Not the hacking}-either.- These were scattered \\in the workshop.(Expanded- being guys with!\\
    {\tiny $\bullet$} \textbf{Not the hacking} of private material of elected\\ officials, e.g. emails, even if the!\\
    {\tiny $\bullet$} \textbf{Not the hacking} has happened yet!!!!!!!!!!!!!!\\
    {\tiny $\bullet$} \textbf{Not the hacking} rumours have cost him any of\\ his followers, least of all the proprietors of!\\
    {\tiny $\bullet$} \textbf{Not the hacking} group behind the breach of Sony,\\ which has posted the staffer's information online,!\\
    }\\
    
    \hline
    
    \makecell[lt]{
     {\tiny $\bullet$} \textbf{How is our} new technology helping us to do\\ that?  We are using a new technology!\\
     {\tiny $\bullet$} \textbf{How is our} system different from that of \\the United States?  The United States is a!\\
     {\tiny $\bullet$} \textbf{How is our} approach different from that of \\the other major European countries? \\ The European Commission!\\
     {\tiny $\bullet$} \textbf{How is our} country going to be able to \\compete with the rest of the world if we don!\\
     {\tiny $\bullet$} \textbf{How is our} country going to be able to \\compete with China in the future?" he asked.!\\
     {\tiny $\bullet$} \textbf{How is our} work different from that of other\\ organizations?  The work of the Center for!\\
     {\tiny $\bullet$} \textbf{How is our} work different from other research\\ in this area?  We are not the first!\\
     {\tiny $\bullet$} \textbf{How is our} system of government supposed to\\ work?  The reason we have a government is!\\
     {\tiny $\bullet$} \textbf{How is our} system different from the one that\\ was used in the past?  The system!\\
     {\tiny $\bullet$} \textbf{How is our} country supposed to be a beacon of \\hope for the world if we have to look!\\
    } &
    
    \makecell[lt]{
     {\tiny $\bullet$} \textbf{How is our} government going to catch up with\\ the cyber criminals?" he said. "I'm!\\
     {\tiny $\bullet$} \textbf{How is our} society selling humanity on \\slavery? The answers to these questions are also\\ important for us!\\
     {\tiny $\bullet$} \textbf{How is our} minister giving it to you? Isn't\\ it? It's got a bit of!\\
     {\tiny $\bullet$} \textbf{How is our} research different from other \\studies?  This study examined the effects of peer-!\\
     {\tiny $\bullet$} \textbf{How is our} mission different from Seniors'\\ Service Corps (SSC) other than the fact!\\
     {\tiny $\bullet$} \textbf{How is our} challenge different?  The only \\difference is that this challenge is about building\\ an!\\
     {\tiny $\bullet$} \textbf{How is our} nation governed?"  As Obama moved\\ into his second term, he is increasingly!\\
     {\tiny $\bullet$} \textbf{How is our} recommendation different from what\\ more traditional veterinarians do?  We don't\\ believe!\\
     {\tiny $\bullet$} \textbf{How is our} rapid abandonment of critical\\ thinking, knowledge, and values, and the \\subsequent burial of!\\
     {\tiny $\bullet$} \textbf{How is our} education system designed for our \\futures?  We are the children of immigrants,!\\
    } &
    
    \makecell[lt]{
     {\tiny $\bullet$} \textbf{How is our} Internet even even connected with our\\ corporate tracks? Every cell phone on the planet\\ knows!\\
     {\tiny $\bullet$} \textbf{How is our} developer name attached to the icon?\\ Since the Planetside icon is use internally!\\
     {\tiny $\bullet$} \textbf{How is our} food paradise created? Artificial \\chemical fertilizers. So these aren't GMOs, but!\\
     {\tiny $\bullet$} \textbf{How is our} acquisition* worth - BOARD ROLL \\(Least Significant Equivalents)!\\
     {\tiny $\bullet$} \textbf{How is our} transit plan addressing this\\ problem?  Under our old plans, Burlington Buses!\\
     {\tiny $\bullet$} \textbf{How is our} mind different than any other part\\ of the body?"  A Broader View!\\
     {\tiny $\bullet$} \textbf{How is our} campaign working?  Bitcoin launches\\ alongside psychological research showing that \\people pay a lot!\\
     {\tiny $\bullet$} \textbf{How is our} mentioning application related to a \\related method (\#five with two in queue) page such!\\
     {\tiny $\bullet$} \textbf{How is our} having to resort to roundabout \\hypotheticals to argue that Stewart may secretly\\ want!\\
     {\tiny $\bullet$} \textbf{How is our} blood working out for you?" a\\ statewide voter got an outpouring of rename and!\\
    }\\
    \hline
    
\end{tabularx}
\caption{decTest data samples for promptGen task and different temperatures. Bold text is the 3-words prompt context.}
\label{app:table:samples_test_1_promptgen_temp}
\end {table*}

\begin {table*}
\centering
\scriptsize

\begin{tabularx}{\linewidth}{
|>{\hsize=.33\hsize}X|
>{\hsize=.33\hsize}X|
>{\hsize=.34\hsize}X|
}
    \hline
   \textbf{Response set ($p = 0.208$)} & \textbf{Response set ($p = 0.64$)} & \textbf{Response set ($p = 1$)} \\
    \hline

    \makecell[lt]{
    {\tiny $\bullet$} \textbf{So that's the} story of the last few years.  The\\ current political climate is not!\\
    {\tiny $\bullet$} \textbf{So that's the} end of the first part of this \\series. I hope you enjoyed it.!\\
    {\tiny $\bullet$} \textbf{So that's the} first thing I want to say. I'm \\not going to be the guy!\\
    {\tiny $\bullet$} \textbf{So that's the} thing about being a professional.\\ You have to be able to handle the criticism!\\
    {\tiny $\bullet$} \textbf{So that's the} way it is. I don't think there's \\any way to change it!\\
    {\tiny $\bullet$} \textbf{So that's the} problem.  It's not just that the \\government is failing to protect!\\
    {\tiny $\bullet$} \textbf{So that's the} thing about this. It's not just \\about the money. It's about!\\
    {\tiny $\bullet$} \textbf{So that's the} end of the story.  The next step\\ is to create a custom!\\
    {\tiny $\bullet$} \textbf{So that's the} case.  So, what's the problem?\\  Well,!\\
    {\tiny $\bullet$} \textbf{So that's the} first time I've ever seen a real\\ one.  I'm not!\\
    } &
    
    \makecell[lt]{
    {\tiny $\bullet$} \textbf{So that's the} state of the campaign. Now, what\\ I do want to talk about is!\\
    {\tiny $\bullet$} \textbf{So that's the} thing: For as much as I love TLC,\\ it's hard to!\\
    {\tiny $\bullet$} \textbf{So that's the} idea, anyway. The last two seasons\\ have been about doing that. It!\\
    {\tiny $\bullet$} \textbf{So that's the} end of the half-hour segment.  The\\ next half-hour!\\
    {\tiny $\bullet$} \textbf{So that's the} situation we're in," he said. \\ "We're in the!\\
    {\tiny $\bullet$} \textbf{So that's the} thing, I don't know if you know, \\but in general it's!\\
    {\tiny $\bullet$} \textbf{So that's the} difference between the kinds of \\things that people will be talking about on\\ Wednesday,!\\
    {\tiny $\bullet$} \textbf{So that's the} \$2.3 billion.  Here's the issue:\\ You're!\\
    {\tiny $\bullet$} \textbf{So that's the} standard for using memcpy(). It's \\fine to use memc!\\
    {\tiny $\bullet$} \textbf{So that's the} next step, and the next step is to \\try to figure out what's!\\
    } &
    
    \makecell[lt]{
    {\tiny $\bullet$} \textbf{So that's the} first time you want to punch \\somebody, not miss before." The Seahawks would!\\
    {\tiny $\bullet$} \textbf{So that's the} science behind the \\Broadwell-E processors from Intel that Intel\\ launched last fall!\\
    {\tiny $\bullet$} \textbf{So that's the} instinct from other teams,\\ that they're a headache. - Ramsay MacDonald,!\\
    {\tiny $\bullet$} \textbf{So that's the} white whale right there about \\too much debt.  And then what you!\\
    {\tiny $\bullet$} \textbf{So that's the} end of our discussion about the\\ causes. What happens when we look at the!\\
    {\tiny $\bullet$} \textbf{So that's the} cover of inhibition against\\ "chronic" or "adaptive" stimulants!\\
    {\tiny $\bullet$} \textbf{So that's the} way the story goes, but exactly\\ how is cloud providers going to restrict Their!\\
    {\tiny $\bullet$} \textbf{So that's the} beginning, the beginning of the \\show, I guess five minutes."  !\\
    {\tiny $\bullet$} \textbf{So that's the} Indie Mobile Game Week Honoring\\ Winners!!!!!!!!!\\
    {\tiny $\bullet$} \textbf{So that's the} reason I'm writing, that's why\\ you don't understand why people know!\\
    }\\
    
    \hline
    
    \makecell[lt]{
    {\tiny $\bullet$} \textbf{do you listen} to the music?"  "I don't know.\\ I don't listen!\\
    {\tiny $\bullet$} \textbf{do you listen} to them?"  "I do," he said.\\ "I'm not!\\
    {\tiny $\bullet$} \textbf{do you listen} to the voices of the people?"\\  "I do," said the king!\\
    {\tiny $\bullet$} \textbf{do you listen} to the song?"  "I don't know\\. I don't know!\\
    {\tiny $\bullet$} \textbf{do you listen} to the music?"  "I do." \\ "You're not!\\
    {\tiny $\bullet$} \textbf{do you listen} to the news?  I do. I'm a big\\ fan of the!\\
    {\tiny $\bullet$} \textbf{do you listen} to me?"  "Yes, I do."  "I'm!\\
    {\tiny $\bullet$} \textbf{do you listen} to the other side?"  "I don't \\know. I don't!\\
    {\tiny $\bullet$} \textbf{do you listen} to the other side?"  "I do," \\said the boy. "!\\
    {\tiny $\bullet$} \textbf{do you listen} to the news?  No, I don't. \\I don't listen!\\
    } &
    
    \makecell[lt]{
    {\tiny $\bullet$} \textbf{do you listen} to the current draft?  I listen to the \\current draft. I'm!\\
    {\tiny $\bullet$} \textbf{do you listen} to it?"  It's easy to hear the "why?" \\but when!\\
    {\tiny $\bullet$} \textbf{do you listen} to the people that come here?" \\ "No, I'm too busy!\\
    {\tiny $\bullet$} \textbf{do you listen} to the thing?"  "Of course I do. I've \\been reading!\\
    {\tiny $\bullet$} \textbf{do you listen} to those who are opposing it,\\ who want to create a situation in which a!\\
    {\tiny $\bullet$} \textbf{do you listen} to music or watch TV? How often\\ do you cook or clean? How much!\\
    {\tiny $\bullet$} \textbf{do you listen} to them? It's like the first time you \\got into something and it just!\\
    {\tiny $\bullet$} \textbf{do you listen} to your father? We'll leave it to the\\ gods to decide."  !\\
    {\tiny $\bullet$} \textbf{do you listen} to music? I like to listen to music, \\but I don't really know!\\
    {\tiny $\bullet$} \textbf{do you listen} to my story and see if you like it?"  \\"I think you!\\
    } &
    
    \makecell[lt]{
    {\tiny $\bullet$} \textbf{do you listen} to Human Fly?, which YouTuber\\ Nico Perri collaborated on, and Google!\\
    {\tiny $\bullet$} \textbf{do you listen} to the acapella lyrics out of\\ context and express the feeling?" It's!\\
    {\tiny $\bullet$} \textbf{do you listen} to Michael Kiwanuka-Smith who \\writes, "The American Journalism Review discern!\\
    {\tiny $\bullet$} \textbf{do you listen} to my songs as I said," \\Ramckhalter said. "You feel!\\
    {\tiny $\bullet$} \textbf{do you listen} to U.S. 90 night at this time\\ of the year?  !\\
    {\tiny $\bullet$} \textbf{do you listen} to that as well?"  "The \\question was not, 'Who is!\\
    {\tiny $\bullet$} \textbf{do you listen}?" He asks, leaning forward as\\ he woodenly talks to him. "Listen!\\
    {\tiny $\bullet$} \textbf{do you listen} to those books and sway him so\\ much? No. He was deeply brainwashed!\\
    {\tiny $\bullet$} \textbf{do you listen}?' Simon(lol).I feel like i'm\\ in a Kurdish Genocide. I!\\
    {\tiny $\bullet$} \textbf{do you listen} to value authenticated queries\\ from your menu when running count? And if not, \\then!\\
    }\\
    \hline
    
\end{tabularx}
\caption{decTest data samples for promptGen task and different $p$ values (nucleus sampling). Bold text is the 3-words prompt context.}
\label{app:table:samples_test_1_promptgen_topp}
\end {table*}

\begin {table*}
\centering
\scriptsize

\begin{tabularx}{\linewidth}{
|>{\hsize=.33\hsize}X|
>{\hsize=.33\hsize}X|
>{\hsize=.34\hsize}X|
}

    \hline
    \textbf{Response set ($k = 3$)} & \textbf{Response set ($k = 32$)} & \textbf{Response set ($k = 318$)} \\
    \hline

    \makecell[lt]{
    {\tiny $\bullet$} \textbf{I'm kidding. You} can't do that.  "I've been\\ trying to find!\\
    {\tiny $\bullet$} \textbf{I'm kidding. You} can't be serious."  The\\ man was a little surprised.!\\
    {\tiny $\bullet$} \textbf{I'm kidding. You} can't do that, because the\\ game's a little bit different.!\\
    {\tiny $\bullet$} \textbf{I'm kidding. You}'re the only one who can do\\ it, you know. You're!\\
    {\tiny $\bullet$} \textbf{I'm kidding. You} can have it.  "It's not a\\ bad idea,"!\\
    {\tiny $\bullet$} \textbf{I'm kidding. You} don't need to be so\\ serious."  "No, I!\\
    {\tiny $\bullet$} \textbf{I'm kidding. You}'re not. I'm just a little\\ bit of a dickhead.!\\
    {\tiny $\bullet$} \textbf{I'm kidding. You}'re not. You're not. I know.\\ I know. You!\\
    {\tiny $\bullet$} \textbf{I'm kidding. You} can't do it.  "I don't want\\ to be!\\
    {\tiny $\bullet$} \textbf{I'm kidding. You}'re not going to be able to\\ see it, and you'll be!\\
    } &
    
    \makecell[lt]{
    {\tiny $\bullet$} \textbf{I'm kidding. You}'re too nice.  We need someone\\ in the government to talk!\\
    {\tiny $\bullet$} \textbf{I'm kidding. You} just sit there, I'll have the\\ other side do it. What!\\
    {\tiny $\bullet$} \textbf{I'm kidding. You} can be my roommate for the\\ holidays in a few weeks. You don!\\
    {\tiny $\bullet$} \textbf{I'm kidding. You} can get the full version of\\ the file above, and also the original!\\
    {\tiny $\bullet$} \textbf{I'm kidding. You} don't look very strong." \\ "You look as strong as!\\
    {\tiny $\bullet$} \textbf{I'm kidding. You}'re right about the last guy.\\ He's a nice guy. But!\\
    {\tiny $\bullet$} \textbf{I'm kidding. You}'ll be happy to know it's the \\first time that's happened since!\\
    {\tiny $\bullet$} \textbf{I'm kidding. You} will come down with the same\\ problem, or some other sort of problem!\\
    {\tiny $\bullet$} \textbf{I'm kidding. You} might have seen me in a lot\\ of other things. I'm actually!\\
    {\tiny $\bullet$} \textbf{I'm kidding. You} should go and see a doctor.\\  In fact, I'm!\\
    } &
    
    \makecell[lt]{
    {\tiny $\bullet$} \textbf{I'm kidding. You}'re kidding?"  "I'm not."  "Why!\\
    {\tiny $\bullet$} \textbf{I'm kidding. You}'re not."  "What? A dick of the\\ heart?!\\
    {\tiny $\bullet$} \textbf{I'm kidding. You}'re looking at a new version,"\\ said Zilch, who was!\\
    {\tiny $\bullet$} \textbf{I'm kidding. You} know when someone takes to \\the streets to protest? It's common for!\\
    {\tiny $\bullet$} \textbf{I'm kidding. You} are definitely a complete free\\ agent," said Caruthers.  !\\
    {\tiny $\bullet$} \textbf{I'm kidding. You} can have another at first, but\\ don't start just jumping ahead/!\\
    {\tiny $\bullet$} \textbf{I'm kidding. You}'re just a teenager, aren't you?"\\ It ends there, your!\\
    {\tiny $\bullet$} \textbf{I'm kidding. You} were never fully persuaded." \\ "Perfect, I am not,"!\\
    {\tiny $\bullet$} \textbf{I'm kidding. You} are also in a worse case \\scenario for someone who was on \$2500!\\
    {\tiny $\bullet$} \textbf{I'm kidding. You} know…"  "I should have\\ stopped him; I shouldn't!\\
    }\\
    
    \hline
    
    \makecell[lt]{
    {\tiny $\bullet$} \textbf{Where did he} go?" I ask, looking at him.\\ "I'm not sure. He!\\
    {\tiny $\bullet$} \textbf{Where did he} get the idea to do this?  \\He had been working on a book!\\
    {\tiny $\bullet$} \textbf{Where did he} come from?  He was born in \\the city of Karkaros!\\
    {\tiny $\bullet$} \textbf{Where did he} go?" I asked. "I don't know,"\\ she said.  !\\
    {\tiny $\bullet$} \textbf{Where did he} go?"  "I think he went to\\ the hospital," she said.!\\
    {\tiny $\bullet$} \textbf{Where did he} get the idea for the name?  I\\ think it's a combination of!\\
    {\tiny $\bullet$} \textbf{Where did he} get the idea to make a movie \\about the Holocaust? "I had a lot!\\
    {\tiny $\bullet$} \textbf{Where did he} get that idea?  "I was just trying\\ to make a statement,"!\\
    {\tiny $\bullet$} \textbf{Where did he} get that from? He's a very good\\ writer. I don't know what!\\
    {\tiny $\bullet$} \textbf{Where did he} go? Where was he? Where was\\ he? He's gone.  !\\
    } &
    
    \makecell[lt]{
     {\tiny $\bullet$} \textbf{Where did he} come back from?  [The Doctor is\\ sitting in a chair. Amy!\\
    {\tiny $\bullet$} \textbf{Where did he} find the money?" asked a reporter\\ from the BBC. "Is anybody else there!\\
    {\tiny $\bullet$} \textbf{Where did he} grow up?"  But the boy answered,\\ "He always loved to read!\\
    {\tiny $\bullet$} \textbf{Where did he} get that idea?" he asked. "I didn't\\ know. I've never!\\
    {\tiny $\bullet$} \textbf{Where did he} come from?" You're looking for\\ that missing piece. Maybe you're missing the!\\
    {\tiny $\bullet$} \textbf{Where did he} come from? He was, I think, from\\ a small island about midway between!\\
    {\tiny $\bullet$} \textbf{Where did he} come from, to be sure?" he asked,\\ "I know he came from!\\
    {\tiny $\bullet$} \textbf{Where did he} go?  [A little while later] I am\\ about to say this!\\
    {\tiny $\bullet$} \textbf{Where did he} hear about my story? I couldn't \\tell you. He'd only heard of!\\
    {\tiny $\bullet$} \textbf{Where did he} come from?  From a place called \\"the City of the Sun."!\\
    } &
    
    \makecell[lt]{
    {\tiny $\bullet$} \textbf{Where did he} at the time in his day seek the God\\ he worshipped? He said: "!\\
    {\tiny $\bullet$} \textbf{Where did he} earn his master's degree? He is\\ part of a class of doctoral students who!\\
    {\tiny $\bullet$} \textbf{Where did he} learn to play guitar?" I asked,\\ puzzled.  "Before I joined!\\
    {\tiny $\bullet$} \textbf{Where did he} come from?"  "Australia," said \\Peter.  "How could!\\
    {\tiny $\bullet$} \textbf{Where did he} hear this, you might ask? Of \\course, he'd heard of it.!\\
    {\tiny $\bullet$} \textbf{Where did he} go?  He's probably dead – or dead\\ and buried within the walls!\\
    {\tiny $\bullet$} \textbf{Where did he} earn \$150 million on his way \\to a \$5 billion makeover? !\\
    {\tiny $\bullet$} \textbf{Where did he} learn to make his own sticks, \\or for that matter, hang a stick on!\\
    {\tiny $\bullet$} \textbf{Where did he} learn to skate, anyway?  Go \\here and watch this beautiful skater!\\
    {\tiny $\bullet$} \textbf{Where did he} get this idea from? What do\\ you think about it?  I get!\\
    }\\
    \hline
    
\end{tabularx}
\caption{decTest data samples for promptGen task and different $k$ values (Top-k). Bold text is the 3-words prompt context.}
\label{app:table:samples_test_1_promptgen_topk}
\end {table*}

\begin {table*}
\centering
\scriptsize

\begin{tabularx}{\linewidth}{
|>{\hsize=.33\hsize}X|
>{\hsize=.33\hsize}X|
>{\hsize=.34\hsize}X|
}
    \hline
    \textbf{Context} & \textbf{Response set (high content diversity)} & \textbf{Response set (low content diversity)} \\
    \hline
    
    \makecell[Xt]
    {
    \textbf{Sold Out}\\
    Jane wanted to watch a big new action movie.\\
    She had been waiting a long time for it to come\\ out.\\
    When tickets became available she was too busy.\\
    By the time she had a chance to buy some it was\\ sold out.
    } &
    
    \makecell[Xt]
    {
    {\tiny $\bullet$} Jane cried over the fact that she couldn't watch\\ it and just gave up looking for a ticket.\\
    {\tiny $\bullet$} Jane decided to look for a scalper that would sell\\ her the ticket for the movie that she really wanted\\ to see.\\
    {\tiny $\bullet$} Jane thought it was okay since she can still have\\ a chance to watch it once it gets uploaded in video\\ and movie streaming applications.\\
    {\tiny $\bullet$} Jane posted a status on her social media accounts\\ asking her friends for any spare ticket that she is\\ willing to buy.\\
    {\tiny $\bullet$} Jane resorted to contacting her old friend who is\\ working at a huge movie theater hoping she can\\ help her get a ticket.\\
    } &
    
    \makecell[Xt]
    {
        {\tiny $\bullet$} Jane remembered that she has an old friend who\\ is a manager at a big movie theater so she contacted \\that friend in the hopes that she can buy any spare\\ ticket.\\
        {\tiny $\bullet$} Desperate to watch the movie, Jane called her \\friend, who works at a movie theater, asking for a \\ticket to that movie.\\
        {\tiny $\bullet$} Jane recalled that her friend works at a movie \\theater and hoped that she can help get a ticket \\for that movie.\\
        {\tiny $\bullet$} Jane decided to look for her friend who could\\ possibly have access to tickets for that movie \\since that friend currently works at a movie theater.\\
        {\tiny $\bullet$} Jane realized that her friend might have spare\\ tickets since she is a manager of a movie \\theater showing that film.
    }\\

    \hline
    
    \makecell[lt]{
    \textbf{Beavers.}\\
    My friend has some beavers in his backyard.\\
    They come up from the creek by his house.\\
    He invites my over and we watch them.\\
    We take pictures of them and send them to\\ our friends.
    } &
    
    \makecell[lt]{
    {\tiny $\bullet$} They are fascinating animals.\\
    {\tiny $\bullet$} Our friends love getting the pictures.\\
    {\tiny $\bullet$} Sometimes his dogs chase them.\\
    {\tiny $\bullet$} They are building a dam on the creek.\\
    {\tiny $\bullet$} They won't let us get too close to them.
    } &
    
    \makecell[lt]{
    {\tiny $\bullet$} They are busy gathering sticks to make a dam.\\
    {\tiny $\bullet$} The dam they are building is almost complete.\\
    {\tiny $\bullet$} It's fascinating to see their workmanship building\\ a dam.\\
    {\tiny $\bullet$} They are turning the creek into a pond by building\\ a dam.\\
    {\tiny $\bullet$} They all work together with careful engineering to\\ build a dam.
    }\\
    
    \hline
    
\end{tabularx}
\caption{conTest data samples for storyGen task.}
\label{app:table:samples_test_2_storygen}
\end {table*}

\begin {table*}
\centering
\scriptsize

\begin{tabularx}{\linewidth}{
|>{\hsize=.33\hsize}X|
>{\hsize=.33\hsize}X|
>{\hsize=.34\hsize}X|
}
    \hline
    \textbf{Context} & \textbf{Response set (high content diversity)} & \textbf{Response set (low content diversity)} \\
    \hline
    
    \makecell[Xt]
    {
    kill la kill is still going new episode every thursday
    } &
    
    \makecell[Xt]
    {
        {\tiny $\bullet$} That show sucks\\
        {\tiny $\bullet$} OMG I can't wait\\
        {\tiny $\bullet$} I thought they canceled it\\
        {\tiny $\bullet$} What channel is it on\\
        {\tiny $\bullet$} I only watch nature programs on BBC
    } &
    
    \makecell[Xt]
    {
        {\tiny $\bullet$} Lead actor is soooo hot\\
        {\tiny $\bullet$} Did you see the cliffhanger at the end of the season\\
        {\tiny $\bullet$} I've been waiting for it to return for weeks\\
        {\tiny $\bullet$} I'm totally gonna binge watch last season\\
        {\tiny $\bullet$} I just got into this show and can't stop watching
    }\\

    \hline
    
    \makecell[lt]{
    places apple slices in a bowl so they'll stay fresh
    } &
    
    \makecell[lt]{
        {\tiny $\bullet$} Oh boy, I love apples.\\
        {\tiny $\bullet$} I don't need you telling me how to keep things\\ fresh, take a hike.\\
        {\tiny $\bullet$} Girl, you're the fresh one around here.\\
        {\tiny $\bullet$} This post might be better in the life hacks section.\\
        {\tiny $\bullet$} This is actually a useful bit of advice.
    } &
    
    \makecell[lt]{
        {\tiny $\bullet$} I find merit in this input.\\
        {\tiny $\bullet$} That information will serve me well.\\
        {\tiny $\bullet$} Thanks, that's really good to know!\\
        {\tiny $\bullet$} Such knowledge is certainly beneficial.\\
        {\tiny $\bullet$} Wise words, I will heed them.
    }\\
    
    \hline
    
\end{tabularx}
\caption{conTest data samples for respGen task.}
\label{app:table:samples_test_2_respgen}
\end {table*}

\begin {table*}
\centering
\scriptsize

\begin{tabularx}{\linewidth}{
|>{\hsize=.5\hsize}X|
>{\hsize=.5\hsize}X|
}
    \hline
    \textbf{Response set (high content diversity)} & \textbf{Response set (low content diversity)} \\
    \hline

    \makecell[Xt]
    {
        {\tiny $\bullet$} \textbf{Suppose there's an} escape plan we haven't thought of yet.\\
        {\tiny $\bullet$} \textbf{Suppose there's an} omelet that is the most amazing ever.\\
        {\tiny $\bullet$} \textbf{Suppose there's an} airplane ticket that's even cheaper.\\
        {\tiny $\bullet$} \textbf{Suppose there's an} actual deadline for this paper.\\
        {\tiny $\bullet$} \textbf{Suppose there's an} event that we can go to this weekend.
    } &
    
    \makecell[Xt]
    {
        {\tiny $\bullet$} \textbf{Suppose there's an} airline that costs less.\\
        {\tiny $\bullet$} \textbf{Suppose there's an} flight that isn't as expensive.\\
        {\tiny $\bullet$} \textbf{Suppose there's an} air travel fare, but doesn't cost as much.\\
        {\tiny $\bullet$} \textbf{Suppose there's an} way to fly there that is low cost.\\
        {\tiny $\bullet$} \textbf{Suppose there's an} flight going there and it's not a lot of money
    }\\

    \hline

    \makecell[lt]{
        {\tiny $\bullet$} \textbf{Nothing remotely like} eating a big breakfast.\\
        {\tiny $\bullet$} \textbf{Nothing remotely like} dancing with your wife at the wedding.\\
        {\tiny $\bullet$} \textbf{Nothing remotely like} singing Justin Bieber's greatest hits\\
        {\tiny $\bullet$} \textbf{Nothing remotely like} falling down a hill\\
        {\tiny $\bullet$} \textbf{Nothing remotely like} getting yelled at
    } &
    
    \makecell[lt]{
        {\tiny $\bullet$} \textbf{Nothing remotely like} being super full and satisfied.\\
        {\tiny $\bullet$} \textbf{Nothing remotely like} getting to taste many different foods.\\
        {\tiny $\bullet$} \textbf{Nothing remotely like} starting the day off right.\\
        {\tiny $\bullet$} \textbf{Nothing remotely like} doing exactly what I want to do.\\
        {\tiny $\bullet$} \textbf{Nothing remotely like} feeding myself with great food.
    }\\
    
    \hline
    
\end{tabularx}
\caption{conTest data samples for promptGen task. Bold text is the 3-words prompt context.}
\label{app:table:samples_test_2_promptgen}
\end {table*}

\end{document}